\documentclass[11pt]{article}
\usepackage[margin=0.7in]{geometry}
\usepackage{amssymb}
\usepackage{amsmath}
\usepackage{graphicx}
\usepackage[table,pdftex]{xcolor}
\usepackage[colorlinks=true,citecolor=blue,urlcolor=black]{hyperref}
\usepackage{rotating}
\usepackage{array}
\usepackage{multirow}
\usepackage{algorithm2e}
\usepackage[colorinlistoftodos]{todonotes}
\usepackage[normalem]{ulem}
\usepackage{pbox}
\usepackage{color,soul}
\usepackage{lineno}
{
\definecolor{colornotes}{rgb}{0.9,0,0}

\begin{document}

\title{Reconstructing Subject-Specific Effect Maps}
\author{Ender Konukoglu \and Ben Glocker}
 \maketitle
\begin{abstract}
Predictive models allow subject-specific inference when analyzing disease related alterations in neuroimaging data. Given a subject's data, inference can be made at two levels: global, i.e. identifiying condition presence for the subject, and local, i.e. detecting condition effect on each individual measurement extracted from the subject's data. While global inference is widely used, local inference, which can be used to form subject-specific effect maps, is rarely used because existing models often yield noisy detections composed of dispersed isolated islands. 
In this article, we propose a reconstruction method, named RSM, to improve subject-specific detections of predictive modeling approaches and in particular, binary classifiers. 
RSM specifically aims to reduce noise due to sampling error associated with using a finite sample of examples to train classifiers. 
The proposed method is a wrapper-type algorithm that can be used with different binary classifiers in a diagnostic manner, i.e. without information on condition presence. Reconstruction is posed as a Maximum-A-Posteriori problem with a prior model whose parameters are estimated from training data in a classifier-specific fashion. 
Experimental evaluation is performed on synthetically generated data and data from the Alzheimer's Disease Neuroimaging Initiative (ADNI) database.  Results on synthetic data demonstrate that using RSM yields higher detection accuracy compared to using models directly or with bootstrap averaging.  Analyses on the ADNI dataset show that RSM can also improve correlation between subject-specific detections in cortical thickness data and non-imaging markers of Alzheimer's Disease (AD), such as the Mini Mental State Examination Score and Cerebrospinal Fluid amyloid-$\beta$ levels. Further reliability studies on the longitudinal ADNI dataset show improvement on detection reliability when RSM is used. 
\end{abstract}
\section{Introduction}\label{sec:introduction}
Statistical analysis methods for neuroimaging data are instrumental in detecting condition induced structural alterations. Available methods can process high number of measurements with complex spatial correlation and construct \emph{effect maps}, for example, in the form of detailed volumetric~\cite{ashburner2001voxel} or surface-based~\cite{greve2011absolute,fischl2012freesurfer} maps that highlight changes statistically related to the condition. Effect maps are often used at the population level, where at each measurement, maps indicate statistical relationship between the condition and measurement across the entire population. Either a group analysis technique~\cite{ashburner2001voxel,krishnan2011partial,worsley1997characterizing} or a machine learning based predictive model~\cite{arbabshirani2016single,gaonkar2013analytic,mwangi2014review,rahim2015integrating,ganz2015relevant} is used to compare two cohorts of subjects, one showing the condition of interest and the other not, and estimate relationships. Population-wide effect maps constructed with existing methods have already provided valuable information on anatomical footprints of various diseases, e.g.~\cite{thompson2001cortical, rosas2002regional, burton2004cerebral}, lifestyle choices, e.g.~\cite{garrido1993cortical, miller2003effects, kanai2011structural}, as well as genetics and inherited traits, e.g.~\cite{watkins2002mri, peper2007genetic, thompson2001genetic}.

Information provided in population-wide effect maps is useful, however, not \emph{subject-specific}. When we consider a measurement extracted from a specific subject, population-wide effect maps do not tell us whether the measurement shows disease effect. Therefore, possible analyses on the extracted measurements are limited to population-wide questions. In order to perform subject-specific analyses, methods that can detect subject-specific effects and construct corresponding maps are needed. Furthermore, methods that can do this diagnostically, i.e. without having information on the presence of the condition for the subject, would be highly desirable. Constructing subject-specific effect maps in a diagnostic manner can have multiple applications. In clinical and neuroscience research, subject-specific detections can be used for stratification and identification of subpopulations~\cite{iqbal2005subgroups}. In engineering research, machine learning tools are often ``black-box'' components.  Subject-specific maps can facilitate model improvement by allowing to analyze cases where methods fail.  Lastly, for clinical practice, subject-specific detections can help in diagnosis and grading. 

There has been previous attempts to detect subject-specific effects by extending group analysis techniques, particularly using one-vs-all analysis~\cite{Maumet:2013ha,Maumet:2016kv}. This avenue is promising as the theory developed in group analysis can be applied. The main drawback, however, is the difficulty in applying this approach diagnostically, as condition information for the subject is needed in the analysis. 

\begin{figure}[!htb]
\begin{tabular}{cc}
  \includegraphics[width=0.48\linewidth]{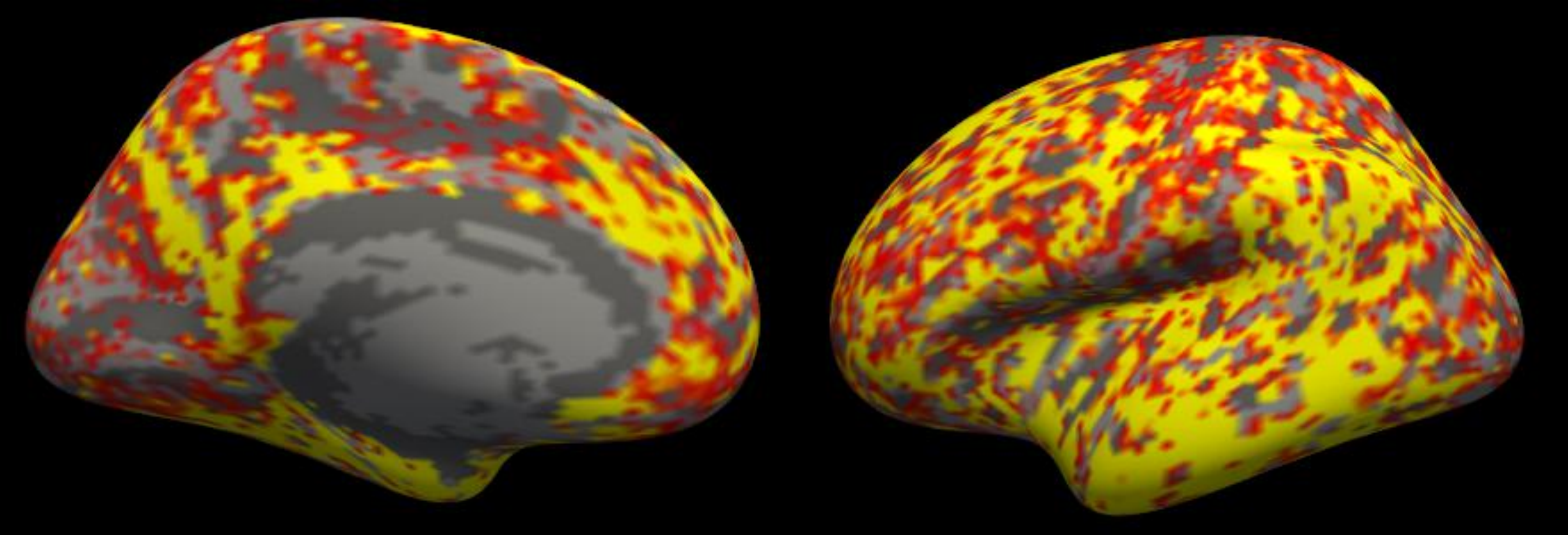} &
  \includegraphics[width=0.48\linewidth]{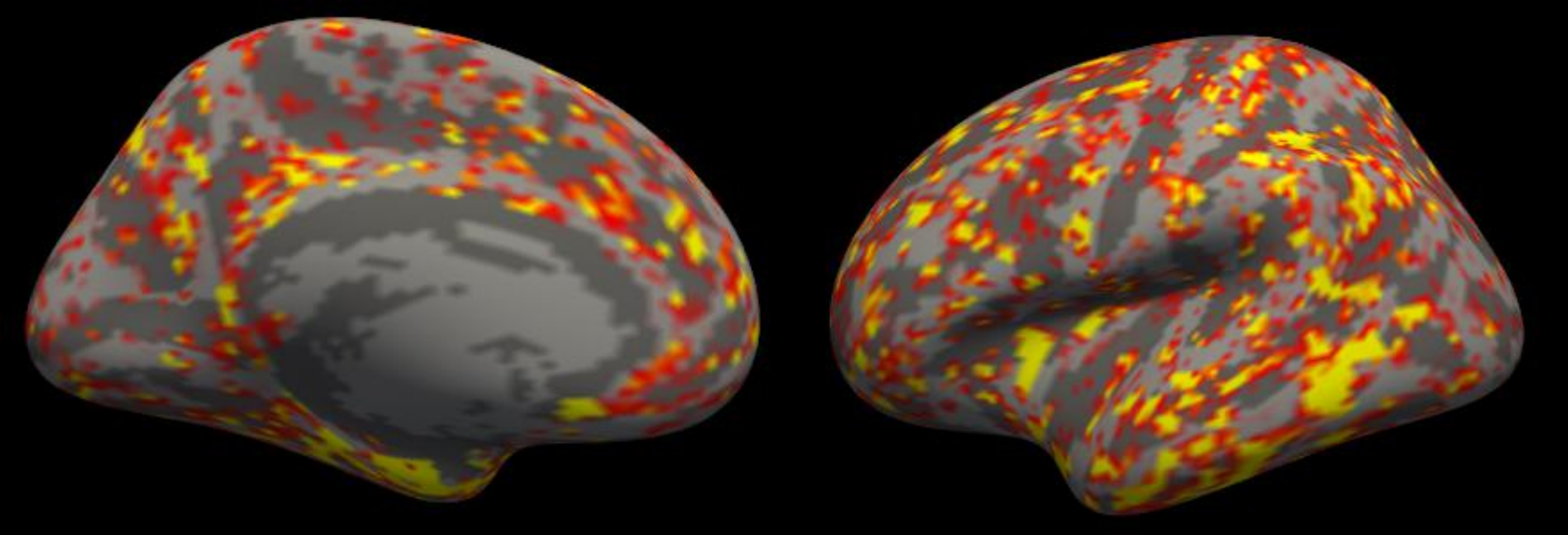} \\
Element-wise GMM & SVM\\
\includegraphics[width=0.48\linewidth]{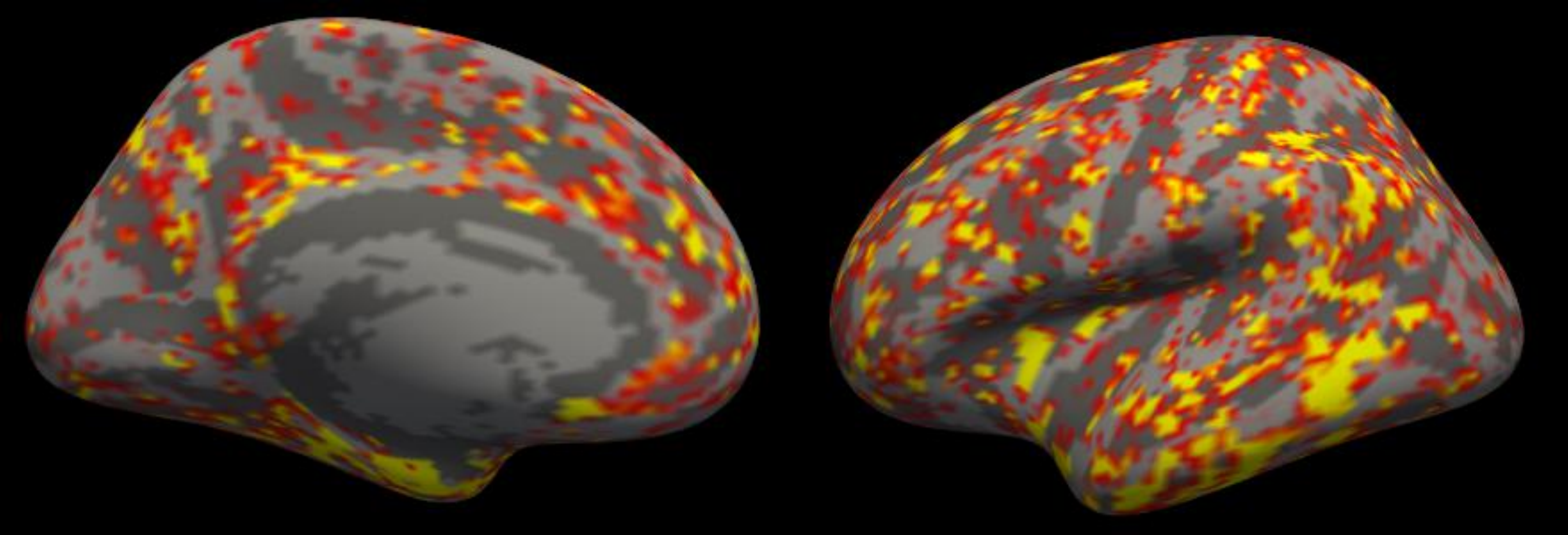} &
\includegraphics[width=0.48\linewidth]{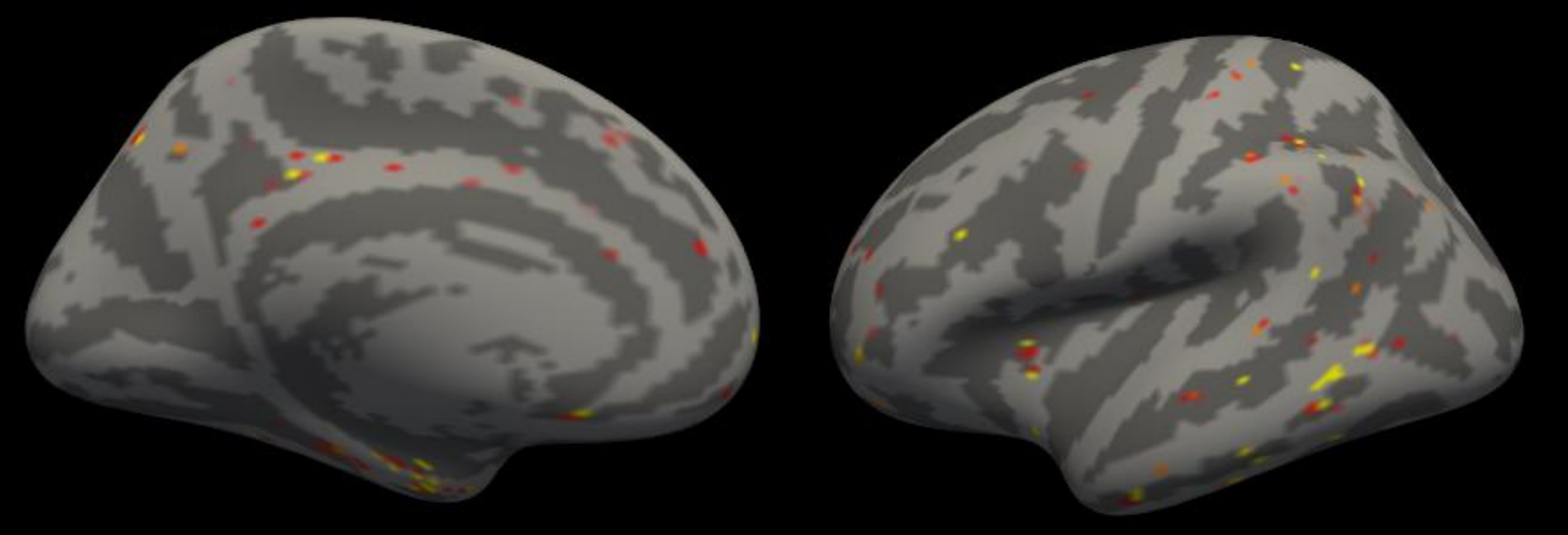} \\
Logistic regression with $L_2$ regularization & Logistic regression with $L_1$ regularization\\
\end{tabular}
\caption{\label{fig:1} \textcolor{black}{{\bf Subject-specific effect maps of Alzheimer's Disease (AD)} extracted from cortical thickness map of a patient with AD using different binary classifiers. We used the ADNI dataset to train the binary classifiers to distinguish between AD patients from healthy elderly using cortical thickness maps of the left-hemisphere extracted using Freesurfer from T1-weighted Magnetic Resonance Images (MRI) (further details on this dataset and the experiments are provided in Section~\ref{sec:experiments}). In the figure, we show subject-specific AD-effect maps for a case subject who was not in the training set. Regions highlighted with yellow are locations where algorithms suggest condition effects with yellow indicating highest degree and red lower. No thresholding is applied and for visualization the same colormap is used for all. Underlying is the inflated left-hemisphere surface with sulci and gyri indicated with different gray levels.  Maps have numerous isolated islands highlighting areas that are not always associated with AD in the literature. However, they also highlight areas that are known to be associated with AD, such as the medial temporal lobe or entorhinal cortex.  This is promising since it suggests that detections have both false and true positives.  A reconstruction method that can suppress the former and highlight the latter would yield cleaner and potentially more useful subject-specific maps.}}
\end{figure}

The main approach for detecting subject-specific effects in a diagnostic fashion is predictive modeling, in particular linear binary classifiers. When trained binary classifiers are applied to a new subject data, algorithms can readily output subject-specific effect maps without algorithmic modification. Despite their availability, these methods are rarely used for this purpose in practice. We believe one of the main reasons for this is that resulting subject-specific effect maps are often ``noisy''. Detections form isolated small islands and can be dispersed to areas that may not be involved in the condition. We illustrate this with an example in Figure~\ref{fig:1} in the context of Alzheimer's disease. Alleviating the noise problem can facilitate subject-specific analyses of neuroimaging data. 

In this article, we present a reconstruction method, named RSM ({\bf R}econstruction {\bf S}ubject-specific effect {\bf M}aps), for improving subject-specific detections of binary classifiers.  
\textcolor{black}{A main source of noise in subject-specific detections is sampling error associated with using finite training sets to train classifiers.  The proposed method reduces this noise by using a Bayesian formulation with a prior probability model formulated as a Markov Random Field (MRF), whose parameters are estimated from training data, and solving a Maximum-A-Posteriori problem. }
RSM is a generic wrapper-type algorithm and can be used with various binary classifiers.  We demonstrate RSM's use with four different models: element-wise Gaussian mixture models (ew-GMM), Support Vector Machines (SVM)~\cite{cortes1995support}, Logistic Regression with $L_2$ and $L_1$ regularization (LR $L_2$ and LR $L_1$). 

We focus on spatial maps of image-based measurements where local measurements are extracted densely at multiple points from the brain.  Examples of such maps are voxel-wise gray matter density~\cite{ashburner2001voxel} and surface-based cortical thickness maps~\cite{fischl2012freesurfer}.  RSM can also be applied to other types of measurements, such as volumes of multiple anatomical structures, but it is especially designed for high-dimensional measurements with spatial context and makes use of the associated correlation structure.  Although our interest is in neuroimaging, the method is not specific to the brain and can be used with other anatomical structures. 

We first describe the proposed method in Section~\ref{sec:method} and then evaluate it in Section~\ref{sec:experiments}.  We performed evaluations both with synthetically generated data, where ground truth information is available, and data from the Alzheimer's Disease Neuroimaging Initiative (ADNI), where the goal is to detect structural alterations due to Alzheimer's Disease (AD).  We present quantitative results focusing on improvements in detection accuracy due to RSM.  On the ADNI dataset, we present a correlation and a reliability study.  The former analyzes correlation between subject-specific detections and auxiliary measures such as Mini Mental State Examination scores (MMSE) and Cerebrospinal Fluid amyloid-$\beta$ (CSF a-$\beta$) measurements.  The latter evaluates reliability of detections in the longitudinal setting.  In both, we focus on the benefits of the proposed method by comparing detections obtained with and without using RSM for the different classifiers. In order to provide a bench-mark we also compare detections with an outlier detection method. We conclude with a discussion in Section~\ref{sec:conclusions}.
\section{Method}\label{sec:method}
The proposed method is a statistical technique to analyze measurements across individuals. In the following we will assume that measurements extracted from different individuals are spatially normalized, which means they are aligned with a common template and corresponding measurements can be directly compared. Such  normalization can be achieved, for instance, using publicly available tools, such as SPM and Freesurfer~\footnote{see \url{http://www.fil.ion.ucl.ac.uk/spm/} or \url{https://freesurfer.net}}.
\subsection{Subject-specific effect maps}\label{sec:subject_maps}
\textcolor{black}{We denote with vector $\mathbf{f} = \left[f_1, \dots, f_d \right] \in \mathbb{R}^d$ a set of measurements extracted from a subject's image. Some examples for $\mathbf{f}$ that are widely used in neuroimaging studies are cortical thickness and gray matter density maps, where each $f_j$ correspond to thickness at a specific location on the cortex and portion of gray matter at a voxel, respectively. We further define $N(j)$ as the set of anatomical locations neighboring the location $j$ where $f_j$ is extracted from. For surface maps, $N(j)$ is the set of vertices that share a face with the vertex $j$ and for volumetric maps it is the set of neighboring grid points based on a pre-defined image neighborhood, such as 6 or 26 neighborhood in 3D.  We further denote the presence or absence of a condition of interest (COI) with a binary variable $y\in\left\{ 0,1\right\}$ (0: absence and 1: presence). When the COI is a disease, then $y$ corresponds to the diagnostic label. $y$ can also correspond to any genotypic or phenotypic variation. }

\textcolor{black}{Binary classifiers are parametric mappings that go from measurements to labels, whose parameters are determined using a training set $D_N = \{\mathbf{f}_n, y_n\}_{n=1}^N$, which consists of both cases (samples with $y=1$) and controls (samples with $y=0$).  Once trained, classifiers can be applied on data coming from a new subject to predict the presence or absence of a COI at subject-specific level.  During this application, each measurement makes a contribution towards a prediction within the trained model.  This contribution can be extracted easily from most models and is the main indicator of condition effect on the measurement, i.e. high contribution towards predicting presence of COI suggests condition effect on the measurement.  Subject-specific effect maps can be constructed by putting contributions of different measurements together.}  

\textcolor{black}{For example, element-wise predictive models compute classification probability $p(y=1 | f_j)$ for each measurement independently. These probabilities can be directly used to construct subject-specific effect maps. Element-wise GMM maps shown in Figure~\ref{fig:1} are constructed this way. On the other hand, multivariate linear models, such as linear SVM and Logistic Regression, use coefficient vectors $\mathbf{w} \in R^d$ for predictions (i.e., one coefficient per measurement) and can compute a subject specific effect map as $\mathbf{w}\circ\mathbf{f} = \left[w_1f_1, \dots, w_df_d\right]$, where $\circ$ is the Hadamard product and $w_jf_j$ is the contribution of the $j^{th}$ measurement towards predicting presence of COI. SVM and Logistic Regression maps in Figure~\ref{fig:1} are such $\mathbf{w}\circ \mathbf{f}$ maps.  Note that, maps coming from different models are not necessarily identical as the computed contributions depend on how the model parameters are determined, e.g. regularization during training would affect $\mathbf{w}$ and therefore $\mathbf{w}\circ\mathbf{f}$.  }
\subsection{\textcolor{black}{Noise in subject-specific effect maps}}\label{sec:noise}
As we mentioned earlier, subject-specific effect maps are often very noisy and this limits their use in practice.  
The main sources of noise are imprecisions in measurements and sampling error associated with the training set.
The former can be mostly attributed to imaging noise, errors in measurement algorithms and inaccuracies in spatial normalization. These issues are mostly addressed by the measurement algorithms. Widely used algorithms are quite robust to imaging noise and can minimize associated measurement imprecisions and algorithmic errors, in the worst case through post-processing quality inspection. Reliability studies of popular algorithms show that measurement imprecision after inspection are indeed minimal \cite{jovicich2009mri,tustison2014large}. 

\begin{figure}
\includegraphics[width=\linewidth]{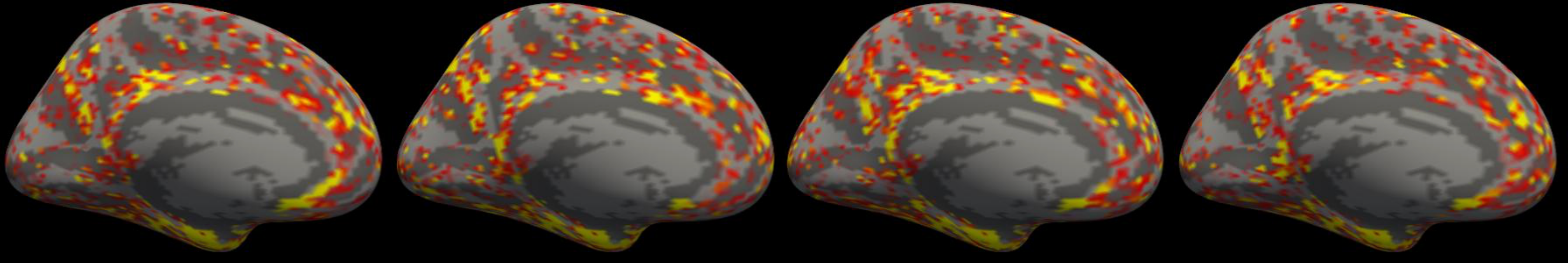}
\caption{\label{fig:2}{\bf Variation of subject-specific effect maps with changing training datasets} is illustrated. Maps shown above are computed for the same case subject diagnosed with AD using four different linear SVMs, each trained with different bootstrap samples of the same training set. Maps show $w_jf_j$ as described in text. In the colormap, yellow is the highest followed by red and no color means value of 0. Highlighted areas have some common regions across the maps, e.g. entorhinal cortex, but also show large variation, in particular in smaller islands. This variability is related to sampling error.}
\end{figure}

Noise due to sampling error, on the other hand, is often not addressed. The training set is effectively a subset sampled from the entire population.  Therefore, there is an error associated with this sampling.  More precisely, the same classification model when trained with different training sets, sampled from the population, can yield different subject-specific effect maps for the same test sample.  In Figure~\ref{fig:2}, we illustrate this on an example AD-effect map computed on a case subject from the ADNI dataset.  The maps are computed using linear SVM trained four times with different bootstrap samples of the same training dataset.  We notice that certain regions are highlighted in all four maps while a large set of regions show variability across the maps. 

One approach to address this variability is to average multiple subject-specific maps computed using different training sets. In the absence of multiple training sets, which is most often the case, one can use multiple bootstrap samples drawn from the training set for the same purpose (cf. ensemble methods such as Random Forests \cite{Breiman2001}). This simple approach reduces noise to some extent as we will show in Section~\ref{sec:experiments} but it can be substantially improved by making better use of the training set.  The method proposed here, RSM, goes in this direction and addresses variability in training set with a dedicated model. 
\subsection{RSM}\label{sec:rsm}
The key intuition behind RSM is to view subject-specific effect maps as noisy observations of underlying true effect maps. Based on this view, we model the reconstruction process as a Maximum-A-Posteriori (MAP) estimation problem where the parameters of the Bayesian model are estimated using bootstrap sampling. 

Let us denote with $C(\cdot)$ a binary classifier that is able to produce subject-specific effect maps and the subject-specific effect map for a new sample with $\hat{\rho}(\mathbf{f})\in\mathbb{R}^d$.  RSM takes as input $\hat{\rho}(\mathbf{f})$ and reconstructs its cleaner version $\rho(\mathbf{f})$ as well as a binary map $\mathbf{q}(\mathbf{f})=[q_1, \dots, q_d]$, in which each element indicates existence of condition effect on the corresponding measurement. The binary map $\mathbf{q}(\mathbf{f})$ is obtained by thresholding $\rho(\mathbf{f})$ with a threshold determined by limiting expected false positive rate. For the sake of notational simplicity, in the remaining, we let go of the dependence on $\mathbf{f}$.

We model the values of $\hat{\rho}$ to be on the real line and the reconstruction model is built with this assumption. Multivariate models, such as SVM and LR, already produce such effect maps where $\hat{\rho} = \mathbf{w}\circ\mathbf{f}$. For element-wise prediction models that produce probabilities at each element, we use the probit function to map the probabilities to $\mathbb{R}$, i.e. $\hat{\rho}_j = \Phi^{-1}(p(y=1 | f_j))$, where $\Phi(\cdot)$ is the cumulative distribution of the standard Gaussian distribution and $\Phi^{-1}$ its inverse. 

We model the effects of classifier variability on the observed effect maps with the following observation model
\begin{equation}
  \nonumber \hat{\rho}_j \triangleq \rho_j + \epsilon_j,\ \epsilon_j \sim \mathcal{N}(0, \sigma_j^2),
\end{equation}
where $\epsilon_j$ is zero mean Gaussian noise with measurement-dependent variance $\sigma_j^2$, and $\mathbf{\rho}=\left[\rho_1, \dots, \rho_d\right]$ is the true continuous effect map, which the method aims to reconstruct. 

Given the observation model, we formulate the reconstruction process as the MAP estimation:
\begin{equation*}
  \mathbf{\rho}^* = \arg\max_{\rho} p(\mathbf{\rho}|\mathbf{\hat{\rho}}) = \arg\max_{\rho} p(\mathbf{\rho})p(\mathbf{\hat{\rho}}|\mathbf{\rho}),
\end{equation*}
where $p(\mathbf{\rho})$ is the prior distribution for $\mathbf{\rho}$. Introducing the conditional independence of the observation model, we can write the MAP estimation as
\begin{equation}
  \label{eqn:map}\mathbf{\rho}^* = \arg\max_{\rho} p(\mathbf{\rho})\prod_{j=1}^d p(\hat{\rho}_j|\rho_j),
\end{equation}
where $p(\hat{\rho}_j | \rho_j) = \exp\{-(\hat{\rho}_j - \rho_j)^2 / 2\sigma_j^2\}/\sqrt{2\pi\sigma_j^2}$.

There are two critical elements in the MAP formulation: the noise variance $\sigma_j^2$ and the prior distribution. $\sigma_j^2$ depends on the test sample under analysis. Ideally, we would estimate it by applying multiple models trained with different training datasets to the test sample. In practice, we approximate it with bootstrap sampling. Specifically, we draw $N_{bs}$ bootstrap samples from $D_N$, retrain $C(\cdot)$ and compute $N_{bs}$ effect maps for the same test subject. For a given test sample, we estimate $\sigma_j^2$ from sampled effect maps with
  \begin{equation}
    \label{eqn:bs_sigma} \sigma_j^2 \approx \frac{1}{N_{bs}}\sum_{r=1}^{N_{bs}} (\hat{\rho}_j^{(r)} - \bar{\rho}_j)^2,\ \bar{\rho}_{j} = \frac{1}{N_{bs}}\sum_{r=1}^{N_{bs}} \hat{\rho}_j^{(r)}, 
  \end{equation}
where $\hat{\rho}_j^{(r)}$  is a bootstrap sample of effect map. Note that the bootstrap averaging approach we described in Section~\ref{sec:noise} uses $\bar{\rho}=\left[\bar{\rho}_1,\dots,\bar{\rho}_d\right]$ as the estimate of the underlying true $\rho$. 

For the prior distribution, we use an MRF model with a unary and a pairwise term:
\begin{equation}
  \label{eqn:prior}p(\mathbf{\rho}) = \frac{1}{Z} \exp\left\{-\frac{1}{2}\left(\sum_{j=1}^d U(\rho_j) +  \sum_{j=1}^d \sum_{k\in N(j)} V(\rho_j, \rho_k)\right)\right\},
\end{equation}
where $Z$ is the normalization constant and $N(j)$ denotes the neighborhood of the $j^{th}$ measurement. We model the unary term to address unreliable measurements and the pairwise term to enforce consistency between measurements extracted from neighboring anatomical locations.  

{Unreliable measurements are those that do not show consistent condition effect within the population. An example would be a measurement whose distribution in the case and control groups overlap. When analyzing a new subject's input map, such measurements may give high $\hat{\rho}_j$ and suggest the presence of the condition. However, these values are most likely due to noise and do not correspond to a true condition effect. In the noise-free effect map, they should be closer to 0 regardless of the observed $\hat{\rho}_j$. To implement this, we model the unary term as follows
\begin{equation}\label{eqn:unary}
U(\rho_j) \triangleq \frac{\rho_j^2}{\varrho_j},
\end{equation}
where $\varrho_j$ is the variance of $\rho_j$ across the population. The underlying idea behind this model is that measurements that show consistent disease effect across the population would yield high $\rho_j$ for cases and low for controls, resulting in high $\varrho_j$. The more consistent the effect, the higher the variance would be. For measurements that do not show consistent effect, we expect the variance to be lower because most cases and controls in the population would get $\rho_j$ values close to 0 with some outliers. As a result, for consistent measurements, $\varrho_j$ would be high and the unary term would allow $\rho_j$ to deviate further from 0. For unreliable measurements, $\varrho_j$ would be low and deviations from 0 would be highly unlikely and penalized.  

We estimate $\varrho_j$ using 5-fold cross validation and bootstrap sampling on the training set. In every fold of the cross validation, the training set $D_N$ is divided into a training part and a test part. 
\textcolor{black}{Multiple models are trained with bootstrap samples of the training part and effect maps for the test parts are computed for each model. Different maps are then averaged over the multiple models to obtain bootstrap average effect maps. Cross-validation assigns a bootstrap average effect map to each sample in the training set and $\varrho_j$ is computed using these average maps}
\begin{equation}
\label{eqn:unary_param}\varrho_j \approx \frac{1}{N} \sum_{n=1}^N (\bar{\rho}_j^{(n)} - \bar{\bar{\rho}}_j)^2,\
\bar{\bar{\rho}}_j=\frac{1}{N}\sum_{n=1}^N \bar{\rho}_j^{(n)}\\
\end{equation}
where $\bar{\rho}_j^{(n)}$ is the bootstrap average effect map of the $n^{th}$ sample in the training set. The cross-validation scheme is illustrated in Figure~\ref{fig:CV}.
\begin{figure}[!htb]
	\begin{center}
	\includegraphics[height=0.14\linewidth]{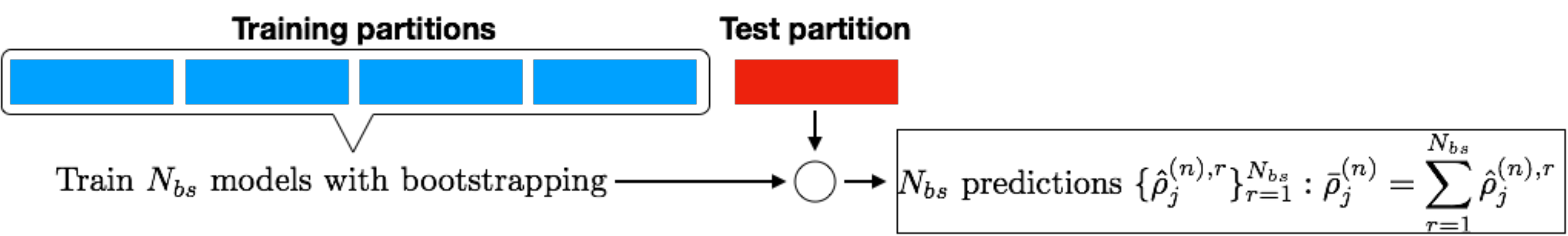}
	\end{center}
	\caption{\label{fig:CV} \textcolor{black}{{\bf Illustration of cross validation to estimate $\varrho_j$ and $\varrho_{jk}$.} $\bar{\rho}_j^{(n)}$ is computed with the cross validation scheme for all samples in $D_N$ and then used to determine the parameters of the prior model using Equations~\ref{eqn:unary_param} and~\ref{eqn:binary_param}.}}
\end{figure}
This procedure requires a cross-validation loop within the training set. For small datasets this may not be feasible. In our experiments, we observed that omitting cross validation by training multiple models on the entire training set, applying the models on the same set and then computing bootstrap average effect maps yield very similar results to the described cross-validation procedure. This option can be used for studies with smaller datasets.

\textcolor{black}{In the reconstruction model we assume consistency of the condition effect across neighboring measurements based on the fact that neighboring measurements might be coming from the same anatomical structure. We use the following pairwise term to enforce this consistency.}
\begin{equation}
  \label{eqn:pairwise}V(\rho_j, \rho_k | \theta_v) \triangleq \lambda \frac{(\rho_j - \rho_k)^2}{\varrho_{jk}},\ k \in N(j)
\end{equation}
where $\lambda$ is a trade-off parameter between the unary and the pairwise term, $N(j)$ is the neighbourhood of the measurement $j$ and $\varrho_{jk}$ is the variance of $\rho_j - \rho_k$ across the population. Low values of $\varrho_{jk}$ indicate high consistency between $j^{th}$ and $k^{th}$ measurements across the population. The pairwise term enforces this consistency when analysing a new sample. High values of $\varrho_{jk}$ indicate lack of consistency and enforcement. Estimation of $\varrho_{jk}$ follows the same procedure as the estimation of $\varrho_j$. We use the same cross-validation scheme illustrated in Figure~\ref{fig:CV} and compute
\begin{equation}
 \label{eqn:binary_param}\varrho_{jk} \approx \frac{1}{N} \sum_{n=1}^N (\bar{\rho}_j^{(n)} -  \bar{\rho}_k^{(n)} - \mu_{jk})^2,\ \mu_{jk} = \frac{1}{N}\sum_{n=1}^N  \bar{\rho}_j^{(n)}-\bar{\rho}_k^{(n)}
\end{equation}
and $\varrho_{jk}=\infty$ when $j$ and $k$ are not neighboring measurements. There are various alternatives for defining the pairwise term. The advantages of the form given in Equation~\ref{eqn:pairwise} are that it is agnostic to the type of measurements and the classifier, and it does not require anatomical segmentation.

So far we assumed the measurements have a spatial structure. However, when there are no spatial relationships between measurements, the pairwise term can simply be ignored. In that case, RSM would reconstruct $\rho$ using only the unary term in Equation~\ref{eqn:unary}. Note that this is not the same thing as not using RSM at all. The unary term would still have an effect on the reconstruction. An example for this case would be if $\mathbf{f}$ is composed of volumes of anatomical structures with no clear spatial relationship between. 

$\varrho_{jk}$ as defined in Equation~\ref{eqn:binary_param} depends on the measurement sites and makes the pairwise term non-stationary. When analyzing a test image, this form allows the model to enforce consistency between two measurements only if they consistently show similar effects for other samples in the population. As a result, the model can respect heterogeneities across measurements that may arise due to anatomical boundaries. One can also consider the stationary alternative $\varrho$, where the variance is estimated as the average of $\varrho_{jk}$ over the pairs of measurement locations as
\begin{equation}\label{eqn:pairwise_sta}
\varrho = \frac{1}{d(d-1)/2} \sum_{j=1}^d\sum_{k>j}^d \varrho_{jk}
\end{equation}
$\varrho$ can be used instead of $\varrho_{jk}$ with RSM however, as we will show in the experiments, its performance is not on par with the nonstationary version. 

\textcolor{black}{The quadratic terms used for defining the unary and binary models yield a Gaussian MRF for $p(\rho)$. These terms are not the only options and one can use different forms to achieve similar effects from the prior model. Our choice of quadratic terms is motivated by the simplicity of the final optimization problem.} Combining the observation model, the unary and the pairwise terms, and taking the logarithm yields the following optimization problem equivalent to the MAP estimation in Equation~\ref{eqn:map}
\begin{equation*}
  \arg\min_{\mathbf{\rho}}\ \sum_{j=1}^d\frac{(\rho_j-\hat{\rho}_j)^2}{2\sigma_j^2} + \frac{1}{2}\sum_{j=1}^d \frac{\rho_j^2}{\varrho_j} + \frac{\lambda}{2}\sum_{j=1}^d\sum_{k\in N(j)} \frac{(\rho_j - \rho_k)^2}{\varrho_{jk}} 
\end{equation*}
Taking the derivatives and setting them to zero yields the following linear system of equations 
\begin{equation}\label{eqn:map_solution}
   \left( \frac{\sigma_j^2}{\varrho_j} + 1\right)\rho_j + \sigma_j^2\lambda\sum_{k\in N(j)}\frac{\rho_j - \rho_k}{\varrho_{jk}} = \hat{\rho}_j,\ j=1,\dots,d.
\end{equation}
The solution of the system of equations given in~\ref{eqn:map_solution} is the reconstructed continuous effect map $\mathbf{\rho}^*$. This system of equations can be solved efficiently using sparse matrix routines in popular linear algebra packages, such as MATLAB and scipy, even for large $d$ when the neighbourhood size $N(j)$ is small for all $j$. \textcolor{black}{Using non-quadratic unary and/or pairwise terms would have necessitated using iterative optimization algorithms, which is arguably more difficult than solving the linear system of equations.}
\subsection{Thresholding}\label{sec:thresholding}
The MAP estimate $\mathbf{\rho}^*$ is a continuous valued map and without a reference value such maps are difficult to interpret and compare across different subjects. In this section, we propose a procedure to generate the binary maps $\mathbf{q}$, which are easier to interpret and can be used for comparisons since they are ``normalized''. In order to determine the binary effect map $\mathbf{q}$, we propose to threshold $\mathbf{\rho}^*$ with a global threshold,
\begin{equation*}
  q_j = \left\{\begin{array}{cl}
  0, & \rho_j^* \leq \tau\\
  1, & \rho_j^* > \tau
  \end{array}\right.
\end{equation*}
There are alternatives for determining the threshold $\tau$. Ideally, one would want to maximize cross-validation detection accuracy on the training dataset. However, this approach would require ground truth condition effects in a set of samples and such information is usually not available nor it is trivial to construct manually. In the absence of ground truth, we assume that the control group in the training dataset is composed of individuals who do not show any condition effect. In other words, $q_j = 0,\ \forall j$ for all control samples. Based on this assumption, we determine the threshold $\tau$ in order to limit the false-positive-rate (FPR) on the control group in the training dataset. Mathematically, we formulate this as
\begin{equation}\label{eqn:threshold}
   \tau = \min t,\ \textrm{such that}\ \frac{1}{dN_0}\sum_{n\in \{y=0\}}\sum_j \delta(\rho_j^{*,(n)} > t) \leq l_{\textrm{FPR}},
\end{equation}
where $\{y=0\}$ denotes the set of control samples in the training dataset, $N_0$ its size, $l_{\textrm{FPR}}$ the desired FPR limit and $\delta(\cdot)$ is the indicator function taking 1 when the argument is true. Optimization given in Equation~\ref{eqn:threshold} is one dimensional and can be solved efficiently with \textcolor{black}{golden section search algorithm~\cite{kiefer1953sequential}}. It aims to determine the minimum value of $t$ that satisfies the $l_{\textrm{FPR}}$, thus avoiding the trivial solution of $t=\max{\rho}$ across the control group and the measurement sites.  

We implement a k-fold cross validation loop to estimate $\tau$ and avoid overfitting. For each fold, the binary classifier is trained and $\varrho_j$, $\varrho_{jk}$ and $\sigma_j$ are computed using the training portion and $\mathbf{\rho}^*$ is computed for the control samples in the test portion. Completing the k-folds yields a $\mathbf{\rho}^*$ map for each control sample in the training dataset. These predictions are used to compute a $\tau$ value based on the desired false positive rate limit. This procedure avoids contamination between estimation of $\tau$ and the other parameters. The cross validation scheme is illustrated in Figure~\ref{fig:CV2}. Estimation of $\left(\varrho_j, \varrho_{jk}\right)$ is performed with an inner cross-validation within the training portion using the scheme described in Section~\ref{sec:rsm}.
\begin{figure}[!htb]
	\begin{center}
		\includegraphics[height=0.14\linewidth]{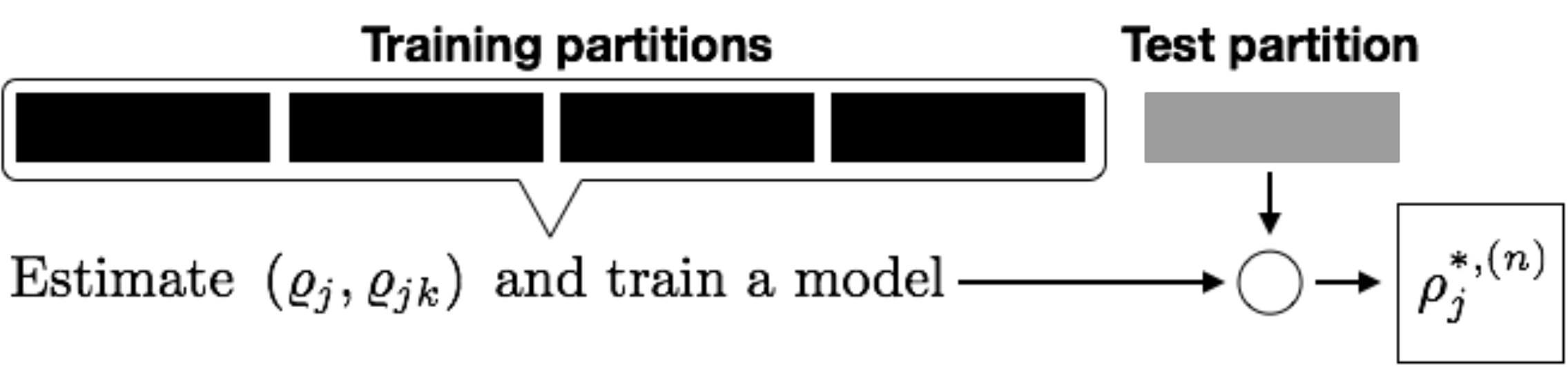}
	\end{center}
	\caption{\label{fig:CV2}\textcolor{black}{{\bf Illustration of cross validation to estimate $\tau$.} $\rho_j^{*,(n)}$ is computed with the cross validation scheme for all control samples in $D_N$ and used to determine $\tau$ by solving the optimization problem given in Equation~\ref{eqn:threshold}. Colors are different from Figure~\ref{fig:CV} on purpose to indicate that these are different cross validation experiments. Cross validation shown in Figure~\ref{fig:CV} is performed in each fold here.}}
\end{figure}

The $l_{\textrm{FPR}}$ threshold limits the FPR over the entire set of measurements therefore, thresholding each element of $\mathbf{\rho}^*$ with $\tau$ avoids the multiple comparisons problem. However, we note two things. First, assuming that subjects in the control group do not have any condition effect is quite strict and makes the threshold $\tau$ conservative. Nonetheless, in the lack of ground truth we opt for this conservative alternative for constructing the binary $\mathbf{q}$ map. Second, $l_{\textrm{FPR}}$ is an expected limit computed on the training set. In test images, actual FPR can exceed this limit due to the randomness. 
\subsection{Tuning parameters:}
There are two tuning parameters of RSM, $\lambda$ and $l_{\textrm{FPR}}$. $\lambda$ controls the strength of the consistency requirement between neighbouring measurements. It is related to the smoothness of the final maps and higher values yield smoother detections. In applications where dense ground truth maps are available, $\lambda$ can be optimized through cross validation. However, the lack of ground truth for disease effect maps makes this impossible and so we let $\lambda$ as a tuning parameter similar to the full-width-half-maximum parameter of smoothing kernels widely used in neuroimaging studies for statistical analysis. 

$l_{\textrm{FPR}}$ on the other hand, controls the amount of false positives a user is willing to accept in the final maps. Increase in the FPR limit yields higher number of detections at the expense of higher false detections. 
\subsection{Implementation details} \label{sec:method_details}
Besides the tuning parameters, all the other parameters of the proposed method are estimated from a training dataset. We explained each estimation procedure in the respective sections. Here, we present remaining details and provide an overall picture of the estimation. 

There are four parameters to estimate: $\varrho_j, \varrho_{jk}, \sigma_j$ and $\tau$. All of them are esimated using cross-validation and bootstrap sampling as explained in Sections~\ref{sec:rsm} and~\ref{sec:thresholding}. Combining both cross-validations yields a nested cross validation that can be infeasible for small datasets. As we have noted previously, when using simple classifiers, such as the ew-GMM, SVM or LR, we empirically observed that opting out from cross validation in the estimation of $\varrho_j$ and $\varrho_{jk}$ is possible without loss of accuracy. For complex classifiers, we would not recommend this because it would increase the chances of overfitting. 

When using bootstrap sampling, we always set the sampling rate to 1 and sample 100 times ($N_{bs}$=100). The number of samples was set empirically and sampling rate of 1 means the size of the sample will be equal to the size of the dataset. When estimating the variances for classification problems, it is often important to keep the original ratio of classes the same in the bootstrap samples. Therefore, we used stratified sampling, where the ratio of cases to controls is the same for each sample and the original dataset.

\section{Experiments}\label{sec:experiments}
We evaluated RSM using synthetically generated data and a cohort of 290 subjects selected from the ADNI dataset. In the experiments with the synthetic dataset, we performed quantitative analysis assessing the detection accuracy of the proposed reconstruction method when used with different binary classifiers. We focused on evaluating improvement in detection accuracy compared to using the binary classifiers directly with naive bootstrap averaging. We also compared detection results with outlier detection as an additional benchmark. For experiments on ADNI, due to lack of ground truth on the condition effect, we performed comparisons with indirect evaluation.

In both experiments, measurements had an underlying spatial structure. In the synthetic data, measurements formed an image and in the ADNI cohort, we used cortical thickness maps extracted using Freesurfer software. Next, we first provide details on the binary classifiers that we used with RSM and the outlier detection method used for additional comparison. Then we present synthetic data experiments and discuss results on the ADNI cohort. 

In all experiments, we implemented RSM using Python 2.7 and performed the experiments on a 64-bit Intel i7-6700K with 32Gb RAM running Debian Linux.
\subsection{Binary classifiers and the outlier detection}\label{sec:bin_classifiers}
We used four different binary classifiers in the experiments: element-wise Gaussian Mixture Models (ew-GMM), Support Vector Machines (SVM), Logistic Regression with $L_2$ regularization (LR $L_2$) and with $L_1$ regularization (LR $L_1$). All these classifiers are widely used and can produce subject-specific effect maps. Here, we briefly explain the way we generated effect maps and implementation details. 
\paragraph{{\it Element-wise Gaussian Mixture Models}} ew-GMM model fits a Gaussian mixture model at every measurement point. The mixture model has two likelihood components, one for cases $p(f_j | y=1) = \mathcal{N}\left(f_j; \mu_1, \sigma_1\right)$ and one for controls $p(f_j|y=0) = \mathcal{N}\left(f_j; \mu_0, \sigma_0\right)$. Using the Bayes rule, posterior distributions can be computed as
\begin{equation}\nonumber
p(y = 1|f_j) = p(f_j | y = 1)p(y=1) / \left(p(f_j|y=1)p(y=1) + p(f_j|y=0)p(y=0)\right),
\end{equation}
where $p(y=1)$ and $p(y=0)$ are the priors. During the training phase, $\mu_i,\sigma_i$ and $p(y=i)$ are estimated using the training dataset as the sample estimates. We used $p(y=1|f_j)$ to compute the subject-specific effect maps after a probit transform, i.e. $\hat{\rho}_j = \Phi^{-1}(p(y=1|f_j))$. ew-GMM model is univariate, meaning $\hat{\rho}_j$ at each point is computed independently from the others. As a result, this model allows for localized interpretations. We used an in-house implementation of ew-GMM in our experiments. 
\paragraph{{\it Support Vector Machines}} We used linear SVM model that performs classification using: $y = \delta(\mathbf{w}^T\mathbf{f} + b > 0)$, where $\delta$ is the indicator function and $\mathbf{w}$ and $b$ are the parameters of the model that are determined by solving the following optimization problem using the training dataset: 
\begin{equation}\nonumber
\min_{\mathbf{w}, b, \xi} \frac{1}{2}\|\mathbf{w}\|_2^2 + \eta \sum_{n=1}^N \xi_n,\ \textrm{such that }y_n(\mathbf{w}^T\mathbf{f}_n + b) \geq 1 - \xi_n,\ \textrm{and }\xi_n\geq 0, n=1,\dots,N,
\end{equation}
where $\eta$ is the regularization parameter. We computed the subject-specific effect map by $\hat{\rho} = \mathbf{w}\circ\mathbf{f} = [w_1f_1,\dots,w_df_d]$. This model is multivariate, meaning $\hat{\rho}_j$ values are computed simultaneously and may influence each other due to the regularization in the model. Therefore, localized interpretations are not trivial. We used the implementation in the scikit-learn package and nested-cross-validation to estimate $\eta$ as the value that maximizes prediction performance. 
\paragraph{{\it Logistic Regression}} We used two different LR models one with $L_2$ and one with $L_1$ regularizations. Similar to SVM, LR model is also linear and performs classification probabilistically with $p(y=1 | \mathbf{f}) = 1 / (1 + \exp(-\mathbf{w}^T\mathbf{f} - b))$. The model parameters $\mathbf{w}$ and $b$ are estimated by maximizing the log-likelihood in the training dataset and when the number of measurements exceeds number of samples, a regularization term $R(\mathbf{w})$ is included to make the problem well-defined: 
\begin{equation}\nonumber
\max_{\mathbf{w}, b} \sum_{n=1}^N y_n\log(p(y=1 | \mathbf{f}_n)) + (1-y_n)\log(p(y=0 | \mathbf{f}_n)) - \eta R(\mathbf{w}),
\end{equation}
where once again $\eta$ is the regularization parameter. We experimented with both $L_2$ and $L_1$ regularization terms, which are commonly used in the literature, i.e. $R(\mathbf{w}) = \|\mathbf{w}\|_2$ and $R(\mathbf{w}) = \|\mathbf{w}\|_1$. Similar to the SVM model, we computed subject-specific effect maps as $\hat{\rho} = \mathbf{w}\circ\mathbf{f}$. LR model is also multivariate and localized interpretations are more difficult than univariate models. We used the implementations in the scikit-learn package and optimized $\eta$ parameter the same way as the SVM model. 
\paragraph{{\it Bootstrap means}} When evaluating RSM, we compared it to bootstrap averaging.  For all the classifiers, we computed bootstrap averaged subject-specific effect maps as described in Section~\ref{sec:subject_maps}, denoted as $\bar{\rho}$ in Equation~\ref{eqn:bs_sigma}.  The bootstrap average had higher accuracy than $\hat{\rho}$ for all the classifiers. In the experiments, we refer to the naive bootstrap averaging approach as WBS.
\paragraph{{\it Thresholding}} Detections of the described binary classifiers are continuous maps. The thresholding technique described in Section~\ref{sec:thresholding} can be used to compute appropriate thresholds and generate binary maps. In the experiments described below, we used this technique to generate binary maps for all classifiers detections whether RSM is used or not. 
\paragraph{{\it Outlier detection}} Although not a condition specific analysis method, outlier detection is another approach that can be used for detecting subject specific alterations.  Techniques for outlier detection estimate normative distributions for the measurements from a population that only consists of individuals not showing the condition, i.e. controls.  For a new subject, the measurements are then compared to the normative distributions and the ones with low likelihoods are determined as outliers.  When using Gaussians for normative distributions, this procedure essentially computes normalized z-scores with outliers having the highest values.  Outlier detection has been used for detecting brain lesions and neurodegenerative changes~\cite{tomasfernandez2015if, VanLeemput:2001jg, Prastawa:2004gk, zeng2016go}.  The main drawback of this approach is its unspecific nature. Outlier detection identifies all measurements that lie outside the respective normative distribution.  The resulting detections are not specific to a condition of interest, hence, cannot be easily used for studying a particular condition.  Pernet et al. also emphasises this lack of specificity in ~\cite{pernet2009brain} even in case of high sensitivity.  

In the experiments, we also used outlier detection to identify subject-specific affected areas for providing another benchmark.  To implement the outlier detection method, we used the control subjects in the training set of each fold to estimate a normative Gaussian distribution and used these distributions to compute likelihood values at each measurement for the test images.  We used the same procedure as in Section~\ref{sec:method} to determine the threshold that limits false positive rates to a user defined bound and obtained corresponding binary subject specific effect maps.
\subsection{Synthetic Data}\label{sec:synth_data}
In order to quantitatively evaluate RSM's contribution to detection accuracy, we generated a synthetic dataset where the ground truth information for the condition effect was available. We generated measurements for 200 samples, where 100  belong to the group with no condition effect, i.e. the control group, and the other 100 to the group with condition effect, i.e. the case group. For each sample, we generated an image of size $100\times 100$ pixels and the pixel intensities are taken as the measurements. Images for the control group contained only stationary noise with spatial covariance and no condition effect. To generate these images, we assigned samples from iid Gaussian noise with zero mean and $\sigma_n=50$ standard deviation to each pixel, and convolved with a Gaussian kernel with standard deviation 2.5 pixels. The convolution yielded correlation between the measurements at neighboring pixels. Example images can be seen in the top row of Figure~\ref{fig:synth_examples}.

We modeled the condition effect as an additive factor on top of the stationary noise. We used two types of effects in order to introduce subject level variation. These two types are shown in the first row of Figure~\ref{fig:synth_examples}. For both, the condition effect was only on the pixels in the white regions in the images with no effect on the measurements in the black areas. The two types of effects shared the central square but differed in the squares at the corners. We generated the images in the case group by first constructing a noisy image similar to the control group and then adding a constant value to the affected pixels. The case group consisted of 50 images per effect type, 100 images in total. Examples images can be seen in the top row of Figure~\ref{fig:synth_examples}. 

The constant value that is added to introduce condition effect is the effect size. We experimented with different effect sizes relative to $\sigma_n$ and will present results for $0.6\sigma_n, \sigma_n, 1.4\sigma_n$ and $2\sigma_n$. We note that different variations of the effect type can also be generated. We chose to generate such a large variation for the sake of illustration.

We tested RSM with all four classifiers listed in Section~\ref{sec:bin_classifiers}. We used 10 randomly shuffled 5-fold cross validation (CV) experiments. Model parameters were estimated with inner 5-fold CV experiments as described in Section~\ref{sec:method}. We defined the pairwise term using the 4-neighbourhood on the image grid. 

\paragraph{Visual results} Some visual results are presented in Figure~\ref{fig:synth_examples}.  In the top row we show six different examples with both the images and the ground truth effect maps.  Effect size for these examples were $1.4\sigma_n$, a relatively low value compared to the noise level.  Next six rows show maps computed with the indicated binary classifiers, using Bootstrap averaging (WBS), and underneath using the same classifiers with RSM using $\lambda=2$ (arbitrarily set).  For each example, we show continuous subject-specific detections, i.e. $\bar{\rho}$ and $\rho^*$, and next to them the thresholded versions, i.e. $\mathbf{q}$ maps. For both $\bar{\rho}$ and $\rho^*$ maps, we computed thresholds by setting $l_{FPR}=0.01$.  In the last row, we show detection results from the outlier detection method, specifically 1 - likelihood values as the continuous maps. 

Results in Figure~\ref{fig:synth_examples} show that RSM ameliorated detections for ew-GMM, SVM and LR $L_2$. Continuous effect maps became cleaner and the thresholded versions captured more of the ground truth effect maps. For LR $L_1$, there were no substantial changes between the detections of the original algorithm and the one that used RSM.  One important point to note is that detections of LR $L_1$ captured less of ground truth compared to the other classifiers whether WBS or RSM was used. 
\begin{figure}[!htb]
  \begin{center}
  	\includegraphics[width=\linewidth]{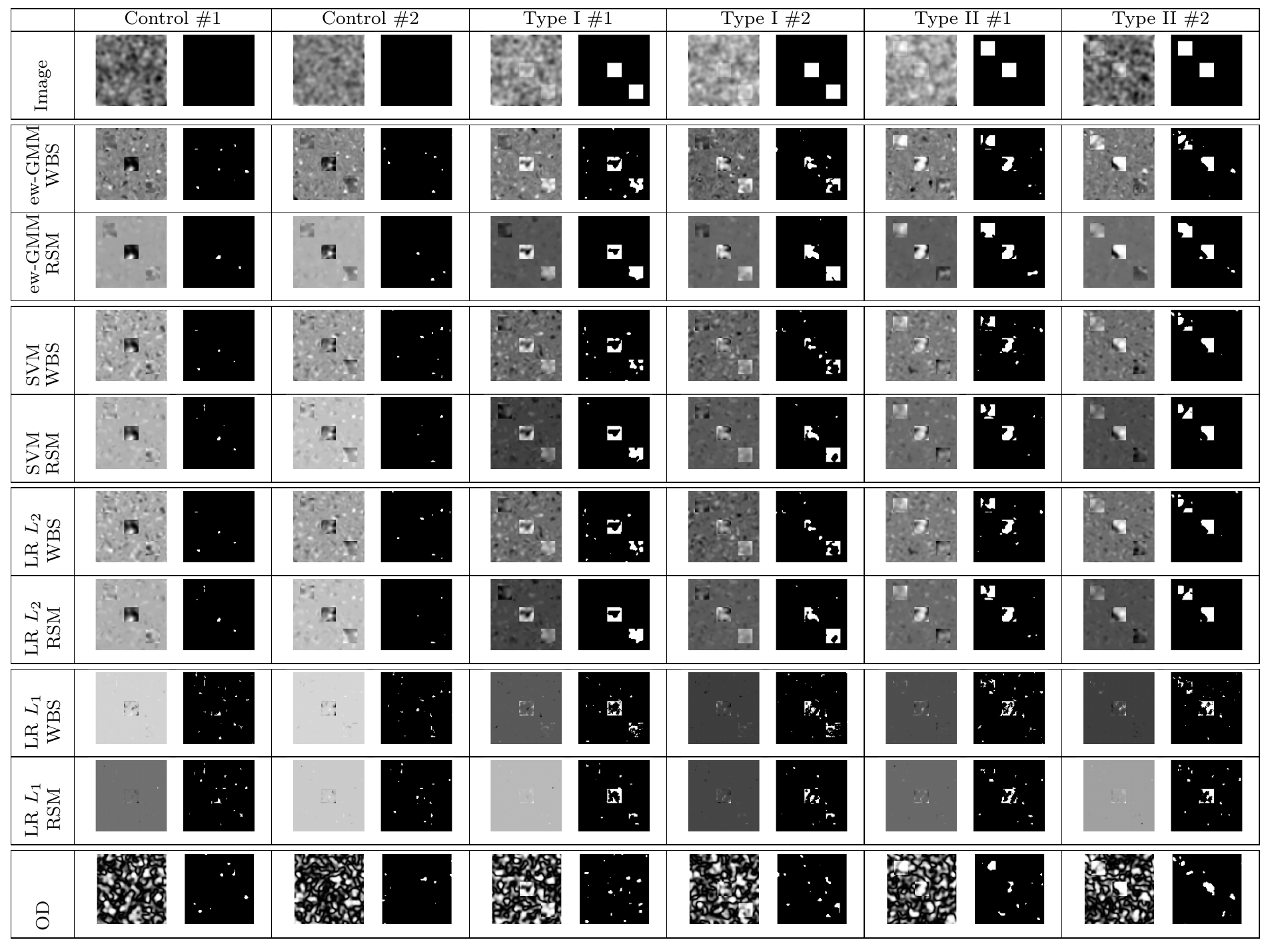}
  \end{center}
\caption{\label{fig:synth_examples}{\bf Visual examples of detections in synthetic dataset.} Figure shows examples of subject-specific detections from the synthetic experiments. The top row show the images and the corresponding ground-truth effect maps for 6 examples. In the next 8 rows, we show detections determined by 4 different classifiers and RSM when used alongside these classifiers. The last row shows detection results using Outlier Detection (OD), i.e. normalised z-scores computed using normative Gaussian distributions.}
\end{figure}

Regarding the outlier detection method, this method did not achieve similar detection accuracy visually as the others. This is not surprising since outlier detection is not condition-specific, therefore, it yielded high number of false positives for all images including controls. Consequently, the thresholding procedure set a high threshold to limit the FPR and resulted in low sensitivity in the final detections. The results presented here are in visual accordance with the results of state-of-the-art outlier detection methods in the literature~\cite{zeng2016go, tomasfernandez2015if}.

In order to illustrate the differences with subject-specific maps, we also performed population-wide analysis. We applied regression analysis (GLM), Random Forests (RFs) and Support Vector Machines (SVMs) to identify population-wide condition effects. We used `statsmodels' package of python~\cite{seabold2010statsmodels} to fit GLM to the entire dataset. We extracted p-value maps and corrected them for multiple comparisons using Bonferroni's method~\cite{bonferroni1935calcolo}. We used the scikit-learn package of python~\cite{scikit-learn} for RF and SVM. We trained RF and SVM using the entire dataset and computed feature importance measures (Gini's criteria~\cite{breiman2001random} for RF and the weights for SVM) at each pixel. Both of these were then converted to p-value maps using permutation testing~\cite{gaonkar2013analytic,good2005permutation}. Both for GLM and other methods, we thresholded the p-value maps at 0.05 level. The population-wide detections of GLM, RF and SVM are shown in Figure~\ref{fig:pop}. As can be seen in the figure, the population-wide effect maps do not provide subject-level information nor information on the effect type variation. Subject-specific effect maps on the other hand, revealed information on both these fronts. 
\begin{figure}[!tb]
  \begin{center}
  	\includegraphics[width=\linewidth]{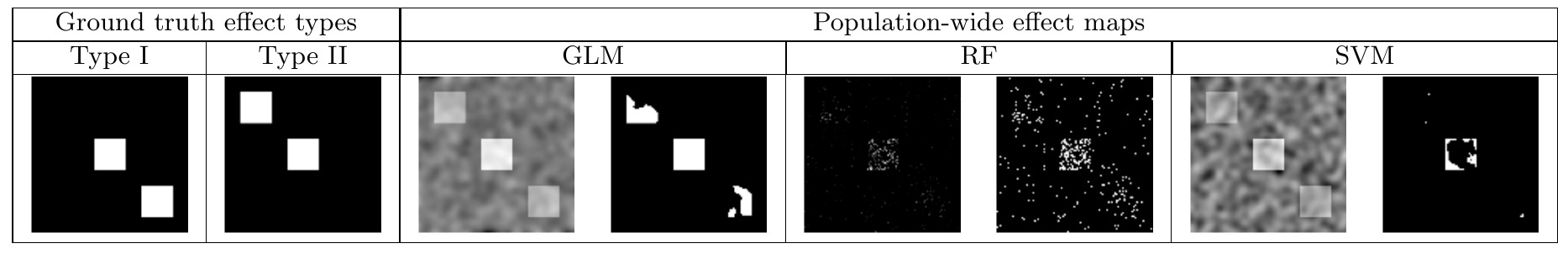}
  \end{center}
  \caption{\label{fig:pop}{\bf Population-wide detection of condition effect with different methods.} Left two are the ground truth for
    the two types of condition effects generated in the synthetic dataset. The remaining 6 are results from GLM, RF and SVM analysis. For each, first image shows the continuous map and the second the thresholded p-value maps at 0.05 level. The GLM map was  corrected for multiple comparisons problem using Bonferroni's method. p-value maps for RF and SVM were obtained via permutation testing but no correction was applied. It is clear that population-wide effect maps do not provide subject-specific information nor can
    they provide variation in effect-type.}
\end{figure}
\paragraph{Quantitative results} In the quantitative analysis, we used Dice's Similarity Coefficient (DSC) and FPR to evaluate the accuracy of the binary subject-specific maps $\mathbf{q}$. We performed evaluations for different $\lambda$ values, FPR thresholds 0.01 and 0.001, and the four different effect sizes: $0.60\sigma_n$, $1.0\sigma_n$, $1.40\sigma_n$ and $2.0\sigma_n$. DSC values are computed only for the case samples and FPR values only for the control samples. Both DSC and FPR were averaged over 10 different random CV experiments. Graphs in Figure~\ref{fig:proposed} plots the DSC and FPR for different $\lambda$ values obtained with different methods. The dashed lines show the scores for the subject-specific maps of binary classifiers using WBS. The solid lines show the scores obtained using RSM with the corresponding binary classifier. 

For ew-GMM, SVM and LR $L_2$, RSM improved the DSC substantially for both FPR limits. DSC values increased with increasing $\lambda$. For LR $L_1$, on the other hand, RSM did not improve DSC for $l_{FPR}=0.01$ and even had a negative effect for $l_{FPR}=0.001$. This behavior can be attributed to a modeling mismatch between RSM and LR $L_1$. The prior in RSM is a Gaussian MRF, i.e. both unary and pairwise terms are quadratic, and the likelihood is also Gaussian. LR $L_1$ on the other hand, relies on $\|w\|_1$ for regularization, which implicitly applies a Laplace distribution as a prior for the contributions of different measurements to the final prediction.

What is perhaps more important was the worse performance of LR $L_1$ in creating subject-specific maps compared to the other methods regardless of using RSM.  We believe, the regularization of LR $L_1$ enforced the method to use as few measurements as possible. This lead to selecting very few measurements that lead to accurate predictions while down-weighing redundant ones. As a result, the method fell short in detecting all affected measurements. This result suggests that LR $L_1$ method may be inappropriate for constructing subject-specific effect maps.

FPR plots show that for $l_{FPR}=0.01$ the method described in Section~\ref{sec:thresholding} successfully limit FPR on the control samples. For $l_{FPR}=0.001$ however, RSM yielded higher FPR when used with ew-GMM and LR $L_1$ than the limit. The violation of the limit was more substantial for lower effect sizes. In absolute value however, the number of false positive detections were minimal on average, e.g. FPR = 0.00150 means 15 falsely identified measurements instead of 5 out of $10^4$. 

The last important point we observed is that at $\lambda=0$ detection results were different compared to using WBS. Applying RSM with $\lambda = 0$, i.e. reconstructing only using the unary term, provided an increase in detection accuracy on its own in some situations. This effect is interesting to note for applications where measurements are not spatially related, and RSM can still be applied using the unary term alone.
\begin{figure}[!tb]
  \begin{center}
  	\includegraphics[width=\linewidth]{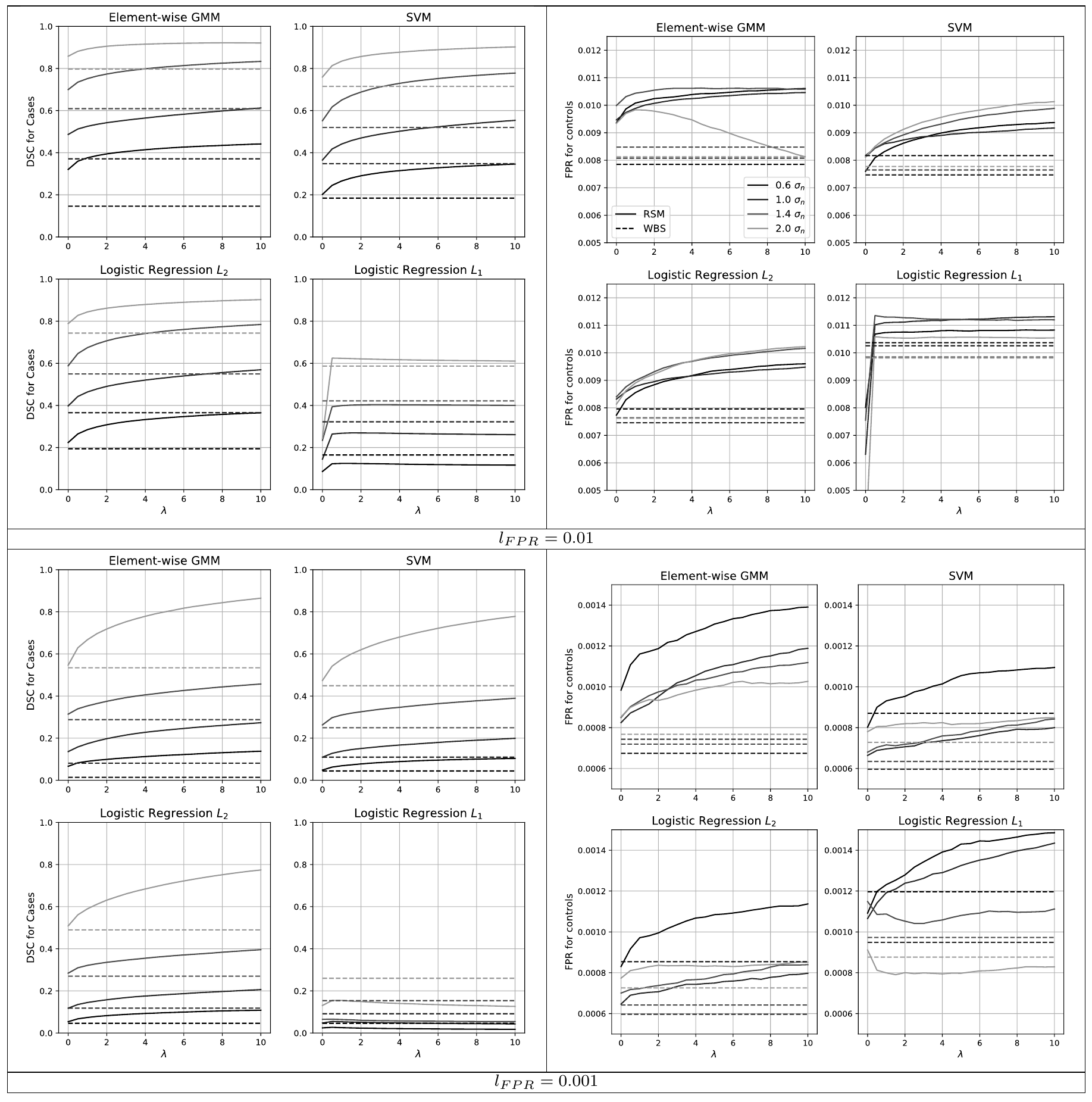}
  \end{center}
  \caption{\label{fig:proposed}{\bf Quantitative results on the synthetic dataset.} Graphs plot DSC and FPR values for subject-specific detections obtained on the synthetic dataset with different binary classifiers using bootstrap averaging (WBS) (dashed lines) and RSM with the same classifiers (solid line). We experimented with different $\lambda$ values, shown in the x-axis, effect sizes, shown in the legend, and $l_{\textrm{FPR}}$. }
\end{figure}

The outlier detection did not perform well in the quantitative assessment. The DCS and FPRS are given in Table~\ref{tab:outlier_synthetic}. We provide them separately not to complicate the plots in Figure~\ref{fig:proposed}. DSC values for $l_{FPR} = 0.01$ were much lower than all the classifiers. Values for $l_{FPR}=0.001$ were similar to the results of LR $L_1$ but much lower than the other classifiers. These low DSC values are not surprising. Outlier detection could not capture condition affected areas due to its unspecific nature. 
\begin{table}
\begin{small}
\begin{center}
  \begin{tabular}{|c|c|c|c|c|}
    \hline
    Effect Size & $0.60\sigma_n$ & $1.0\sigma_n$ & $1.4\sigma_n$ & $2.0\sigma_n$ \\
    \hline
    DCS @ $l_{FPR}=0.01$ & 0.0337 & 0.0860 & 0.2294 & 0.4460 \\
    \hline
    DCS @ $l_{FPR}=0.001$ & 0.0047 & 0.0169 & 0.0740 & 0.2205 \\
    \hline
    FPRS @ $l_{FPR}=0.01$ & 0.0093 & 0.0093 & 0.0094 & 0.0090 \\
    \hline
    FPRS @ $l_{FPR}=0.001$ & 0.0011 & 0.0011 & 0.0001 & 0.0001 \\
    \hline
  \end{tabular}
\end{center}
\end{small}
\caption{\label{tab:outlier_synthetic}{\bf Quantitative results of outlier detection on the synthetic dataset.} DSC and FPRS are given for different effect sizes and $l_{FPR}$. Due to its unspecific nature, outlier detection could not detect accurately the subject-specific affected areas in the synthetic dataset.}
\end{table}

The quantitative assessment provided in this section was based on comparing WBS and the proposed RSM approach on the original images. One can also consider performing the same comparison after filtering the original images to reduce noise. In the supplementary materials we present such an analysis to show that RSM is fundamentally different than filtering the original images.
\subsection{ADNI Dataset}
In the second set of experiments, we applied RSM to analyze the effects of Alzheimer's Disease (AD) on the cortical thickness to detect subject-specific atrophy patterns. We selected a cohort of 290 subjects from the ADNI database, which consisted of 145 patients with AD diagnosis and 145 age and gender matched controls. In addition to age and gender matching, we also made sure that the Freesurfer software package~\cite{fischl2012freesurfer} was able to process the structural images of the selected subjects without problems and the results passed visual quality control. Beyond matching and Freesurfer criteria, the subject selection was random. We used the structural T1-weighted magnetic resonance images of the selected subjects and extracted cortical thickness maps\footnote{Cortical thickness maps consists of gray matter thickness values extracted across the entire cortical mantle and discretized as a triangular surface mesh. Each vertex holds the thickness value of the underlying cortical gray matter. } using the Freesurfer. We aligned the cortical thickness maps of all individuals on a common reference surface mesh defined on the MNI atlas and decimated the number of vertices to 10242 using the Freesurfer for faster computation. We did not apply any surface based smoothing to the thickness maps for the experiments.

We performed 10 randomly shuffled 5-fold CV experiments. As in the previous section, for each fold we estimated the parameters of the proposed method using an independent, inner 5-fold CV loop. For the tuning parameter, we tested different values of $\lambda=\{0,0.1,0.25,0.5,0.75,1,2.5,,5,10\}$ and two different FPR limits $l_{\textrm{FPR}}=\{0.001, 0.01\}$.} In order to define the pairwise term in the restoration formula, we used the triangular surface mesh. We defined the neighbors of a vertex as the set of vertices that shared a mesh triangle with it.
\subsubsection{Correlation with auxiliary markers}
Ground truth for AD affected areas was, unfortunately, not available for in-vivo data. Therefore, we resorted to indirect evaluation strategies. Our main hypothesis is that if the detected regions overlap with truly affected areas, then they should contain condition related information and be statistically related to other auxiliary markers of AD. In our evaluation, we used the Mini Mental State Examination (MMSE) scores and Cerebrospinal Fluid amyloid-$\beta$ (CSF a-$\beta$) levels as the auxiliary markers and computed correlations with the areas of detected regions. In addition, we also computed group differences in size of areas of detected regions between the AD and control groups using t-statistics. 

We detected AD affected measurements for each subject using the binary classifiers with and without RSM in the 5-fold CV setup. Each measurement in a cortical surface map corresponds to a vertex and we considered the number of detected vertices to roughly correspond to the area of the detected region on the cortical mantle. We performed the same comparative study as in the synthetic data experiments. Specifically, we used all the four binary classifiers, i.e. ew-GMM, SVM, LR $L_2$ and LR $L_1$, to identify subject-specific areas using bootstrap averaging and RSM reconstruction. For all methods, we constructed binary maps using the thresholding technique described in Section~\ref{sec:thresholding}. We then computed Pearson's correlation coefficients between the auxiliary measures and the number of detected vertices in the binary maps. The MMSE scores were available for all subjects in our cohort while only 147 subjects had CSF a-$\beta$ measurements. In addition to the classifiers, we used the outlier detection method described in Section~\ref{sec:synth_data} to identify subject-specific abnormal regions and compute correlations with auxiliary markers in the same way. Lastly, the generated binary maps were also used to compute group differences in the number of detected vertices.

In order to provide a reference for the correlation values, we also trained dedicated Random Forest (RF) regressors that predict the auxiliary markers directly from aligned cortical thickness maps. Similar to detection experiments, we performed 10 randomly shuffled 5-fold CV experiments to quantify the RF-based regression accuracy. 

There are three points we would like to note. First, we chose to use the number of identified measurements, or alternatively the area of the affected region, because it is conceptually a very straightforward statistic for quantifying the disease load. Other statistics could also be used for the quantification. Second, we chose to use MMSE and CSF a-$\beta$ as the auxiliary markers, instead of say hippocampus volume, because these markers are not image-based, and therefore, the correlations with the detected areas are less likely to be contaminated by spurious dependencies. Lastly, we note that correlation coefficient does not assess whether a method detects all affected areas for a subject. It assesses whether a method's detection is proportional to the auxiliary measure. If a method consistently underestimates or overestimates affected areas for all subjects, it can still achieve a high correlation coefficient in absolute value. Correlation will be low if detections do not have any statistical relationship. 
\begin{figure}[!htb]
  \begin{center}
    \includegraphics[width=0.95\linewidth]{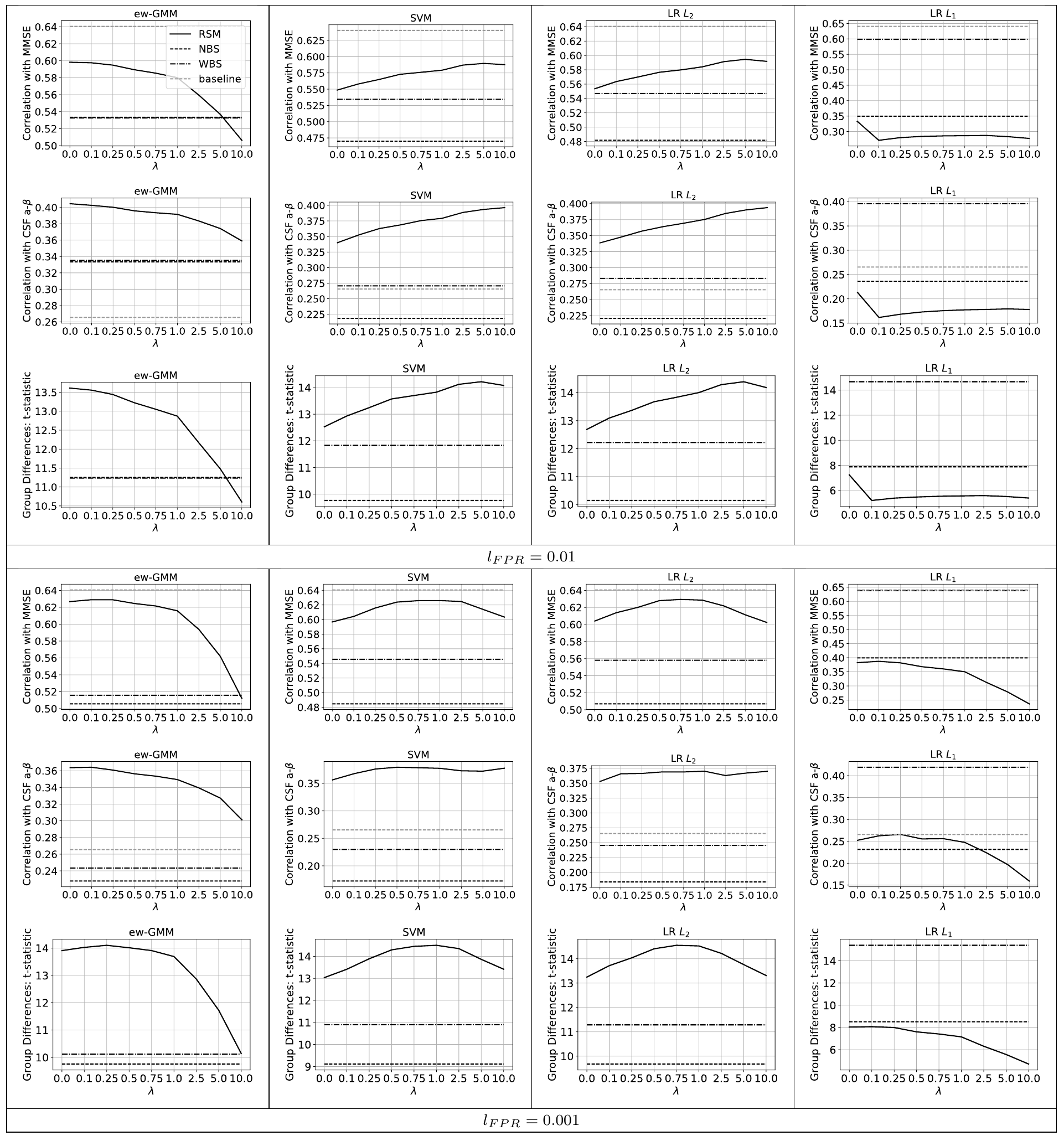}
  \end{center}
  \caption{\label{fig:aux_measures}{\bf Correlation between areas of subject-specific detections and auxiliary measures.} The top group of 12 plots show results for $l_{FPR}=0.01$ and the bottom for $l_{FPR}=0.001$. Each column shows results for one of the four classifiers when used directly (NBS), with bootstrap averaging (WBS) and RSM with different $\lambda$. The top and middle plots in each group show correlation scores between the number of detected vertices and (top) MMSE and (middle) CSF a-$\beta$. The bottom plots show group differences in number of detected vertices between AD and control groups as computed with t-statistic. The reference baselines in MMSE and CSF a-$\beta$ graphs are correlation scores obtained with dedicated Random Forest algorithms trained to predict the measures using cortical thickness maps. We used the same visualization for each plot.}
\end{figure}

Graphs in Figure~\ref{fig:aux_measures} plot the correlation scores for MMSE and CSF a-$\beta$ in absolute value, and group differences between the AD and control groups. In each plot, we show results for detections with no bootstrap averaging nor RSM (indicated with ``NBS'') with bootstrap averaging (indicated with ``WBS'') and using RSM with different $\lambda$ values. For ew-GMM, SVM and LR $L_2$, using RSM lead to a substantial increase in correlation with MMSE compared to both NBS and WBS. Correlation scores using only the areas of the detected regions came close to the correlation score the dedicated RF obtained. For LR $L_1$, bootstrap averaging yielded a large increase over NBS but RSM lead to a decrease. The CSF a-$\beta$ graphs show a similar behavior. Interestingly, correlations of subject-specific maps were higher than that of the dedicated RF. Lastly, for group differences, we observed again a similar behavior. RSM enhanced group differences for ew-GMM, SVM and LR $L_2$ but had adverse effects for LR $L_1$. 

For SVM and LR $L_2$, the highest correlation scores were achieved for $\lambda>0$. However, using very high $\lambda$ values did not help as RSM started to show adverse effects in some situations due to over-smoothing. While SVM and LR $L_2$ benefitted from increasing $\lambda$, ew-GMM achieved highest correlation at $\lambda = 0$, i.e. restoration with only the unary term.

RSM was beneficial for ew-GMM, SVM and LR $L_2$, but it was detrimental to LR $L_1$. This behavior can again be attributed to a mismatch between the modeling assumptions behind RSM, i.e. Gaussian MRF prior, and LR $L_1$, Laplace prior. We also observed that WBS in LR $L_1$ performed very well and achieved very high correlation scores and group differences. This result does not contradict with the lower performance in the synthetic data experiments. As we noted, high correlation or group difference does not necessarily mean the method identifies all subject-specific effects.

Lastly, correlation scores and group differences for the outlier detection were much lower than those of the binary classifiers. For $l_{FPR}=0.01$ the correlation scores in absolute value were 0.20 and 0.04 for MMSE and CSF a-$\beta$, respectively. The values for $l_{FPR}=0.001$ were 0.15 and 0.07. The group differences at $l_{FPR}=0.01$ and $0.001$ were 2.51 and 2.52 respectively. These results were similar to what we observed for the synthetic data. Unspecific nature of outlier detection was not enough to identify condition related alterations at the subject-specific level. 
\subsubsection{Visual results}
\begin{figure}[!htb]
  \begin{center}
    \includegraphics[width=\linewidth]{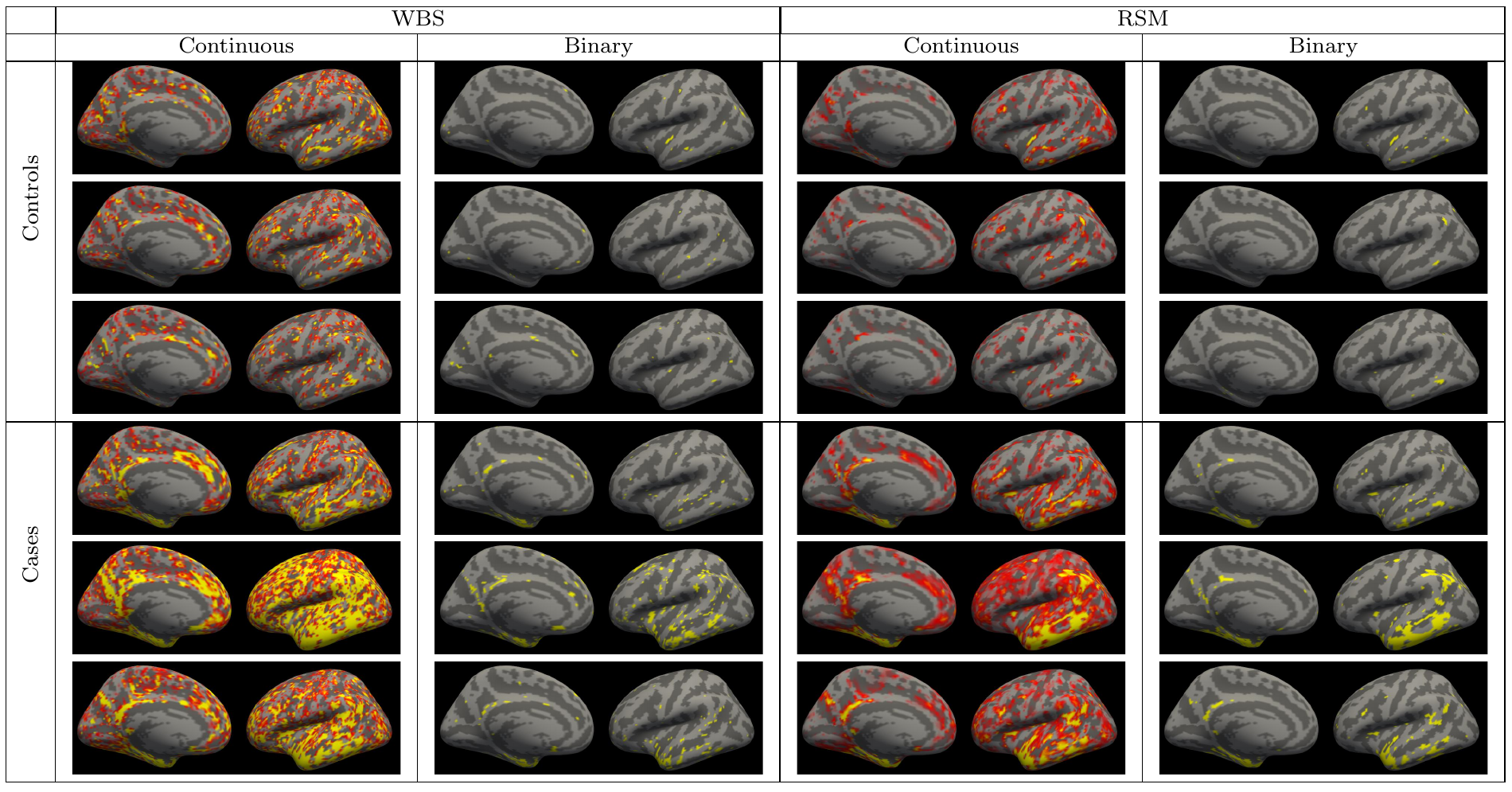}
  \end{center}
  \caption{\label{fig:adni_visual_ewgmm} {\bf Visual examples of subject-specific detections for ADNI using ewGMM.} Results are shown for bootstrap averaging (WBS) and RSM with $\lambda=2$ and $l_{FPR}=0.01$. Continuous maps for WBS are $\bar{\rho}$ maps computed using Equation~\ref{eqn:bs_sigma} and for RSM they are $\rho^*$ computed by solving Equation~\ref{eqn:map_solution}. Binary maps are obtained by thresholding the continuous maps as described in Section~\ref{sec:thresholding}. The same setting and visualization, i.e. colorscale, is also used for all images in Figures~\ref{fig:adni_visual_svm}-\ref{fig:adni_visual_lrl1}}
\end{figure}

\begin{figure}
  \begin{center}
    \includegraphics[width=\linewidth]{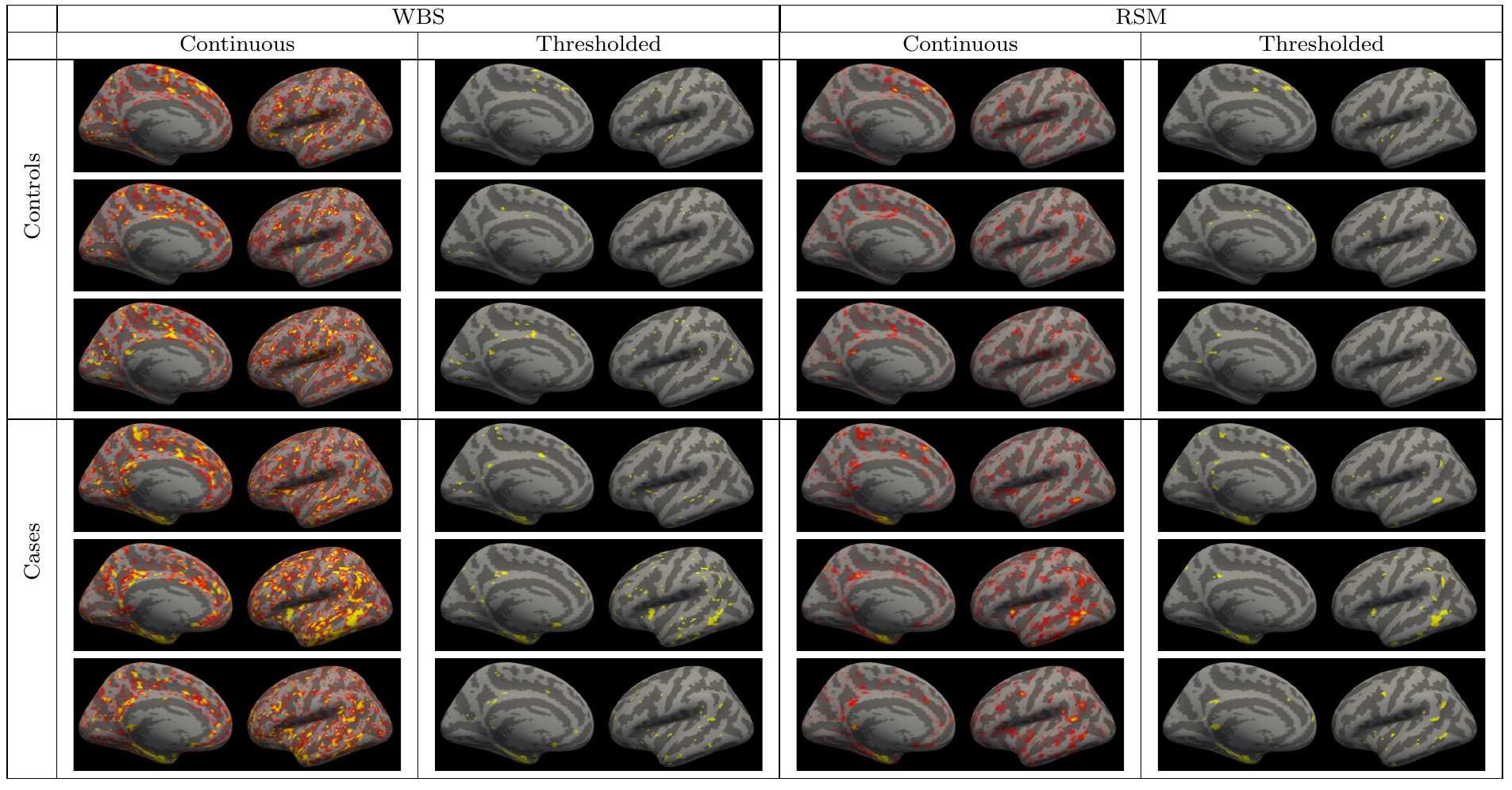}
  \end{center}
  \caption{\label{fig:adni_visual_svm} {\bf Visual examples of subject-specific detections for ADNI using SVM.}}
\end{figure}

\begin{figure}
  \begin{center}
    \includegraphics[width=\linewidth]{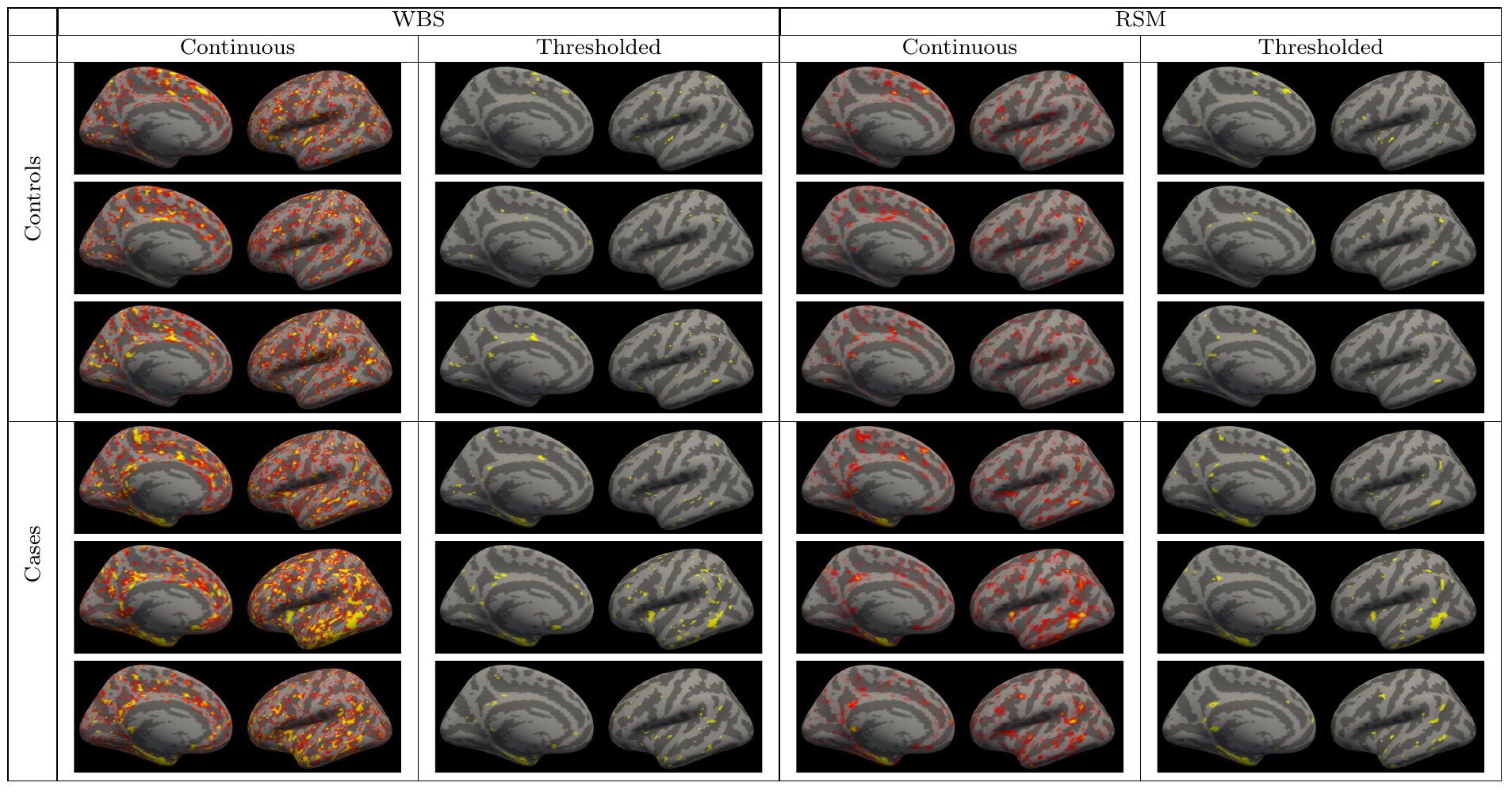}
  \end{center}
  \caption{\label{fig:adni_visual_lrl2} {\bf Visual examples of subject-specific detections for ADNI using LR $L_2$}}
\end{figure}

\begin{figure}
  \begin{center}
    \includegraphics[width=\linewidth]{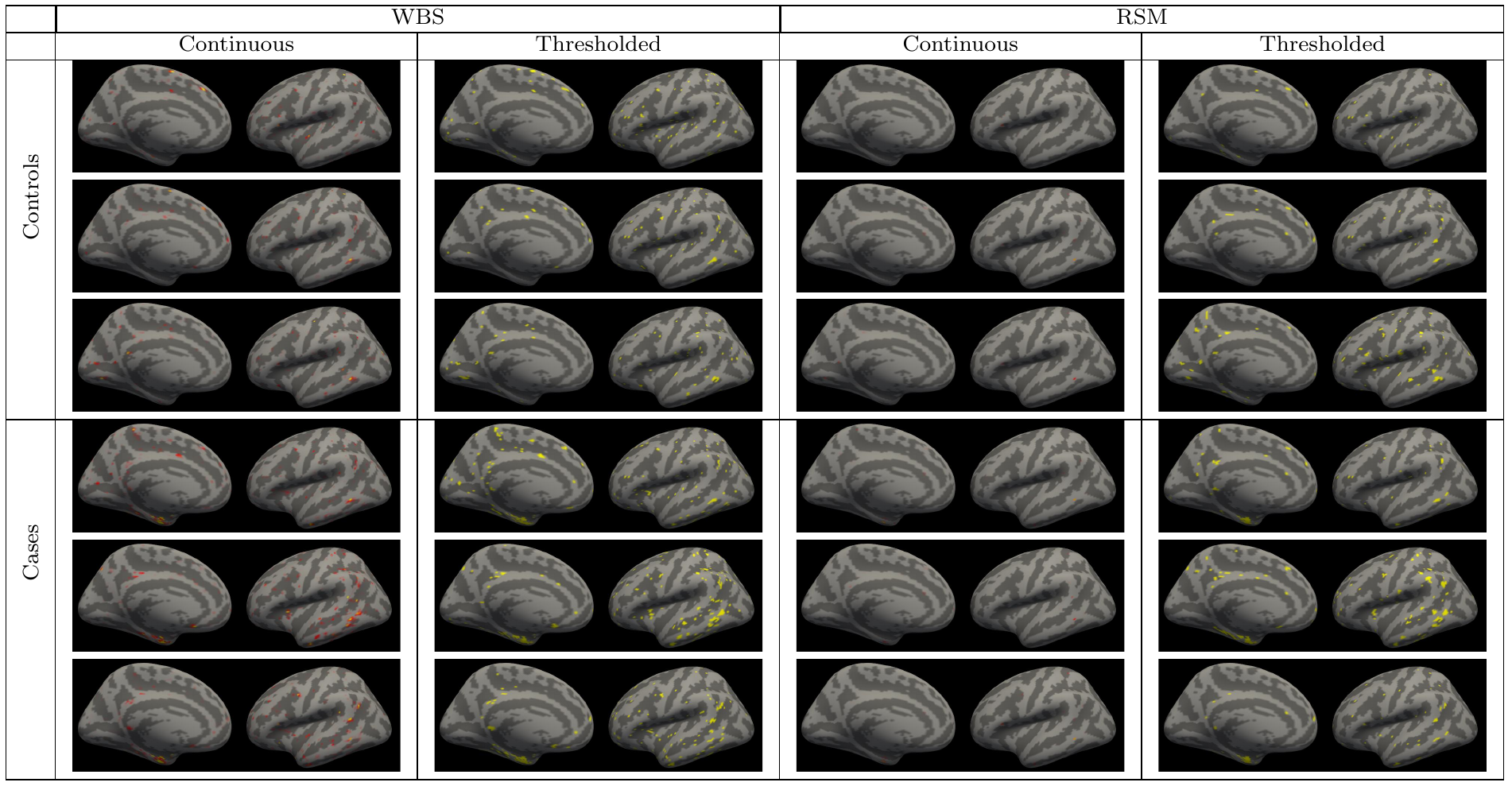}
  \end{center}
  \caption{\label{fig:adni_visual_lrl1} {\bf Visual examples of subject-specific detections for ADNI using LR $L_1$}}
\end{figure}

To complement the quantitative results presented above, we present visual results in this section. Figures~\ref{fig:adni_visual_ewgmm}-~\ref{fig:adni_visual_lrl1} show detection results for three AD subjects and three controls for the different classifiers. For the visual results, we set $\lambda=2$ and $l_{FPR}=0.01$. For all subjects and classifiers, we show the continuous detection maps identified with RSM and bootstrap averaging (indicated with ``WBS'') as well as the corresponding thresholded binary maps. For visualization, all continuous maps are shown in the same color-scale and yellow indicates highest value. We note that interpreting continuous maps is not possible as the absolute values do not have a special meaning. Thresholded maps however can be directly compared between WBS and RSM. The effects of RSM are clear in the thresholded maps of cases. RSM generated maps have fewer but larger contiguous areas than WBS generated maps.

\begin{figure}
  \begin{center}
    \includegraphics[width=\linewidth]{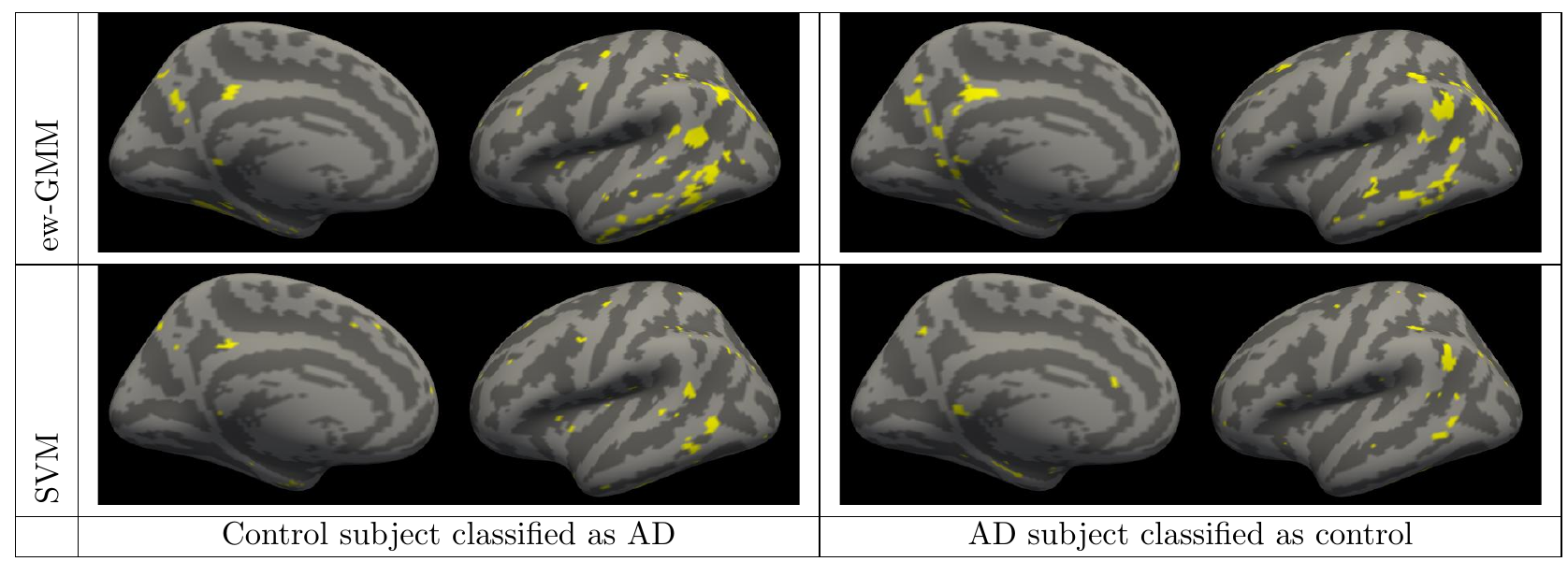}
  \end{center}
  \caption{\label{fig:rf_misclassification}{\bf Explaining misclassification.} Figure shows subject-specific detections for some subjects where a dedicated random forest (RF) misclassified. Detections are obtained using RSM with two different classifiers, i.e. ew-GMM and SVM. Left column shows the detections for a control subjects where the dedicated RF predicted AD. Right column shows the detections for an AD patient where the RF predicted control. Subject-specific maps on the left column shows large number of detections in various areas including temporal lobe and entorhinal cortex, which might explain the AD classification. On the right, fewer such detections might be leading to a control classification. RF classifier's probabilistic outputs (probability of being AD patient) were 0.52 for the subject on the left and 0.40 for the one on the right.}
\end{figure}

In order to analyze the value of subject-specific detections and RSM for understanding decisions of machine learning tools, we trained a Random Forest (RF) classifier to distinguish between AD patients and controls using the cortical thickness maps. We trained and tested the RF in a 5-fold CV setup. In Figure~\ref{fig:rf_misclassification}, we show subject-specific detections of two subjects, one AD and one control, for which the dedicated RF classifier misclassified the subjects. We show detections computed using only ew-GMM and SVM for the sake of simplicity. RF's misclassification can be interpreted easily by analyzing the subject-specific detections. For the control subject, there were many detections around areas that are associated with AD. For the AD subject on the other hand, detections were fewer which might have led the RF to misclassification. 

\begin{figure}
  \begin{center}
    \includegraphics[width=\linewidth]{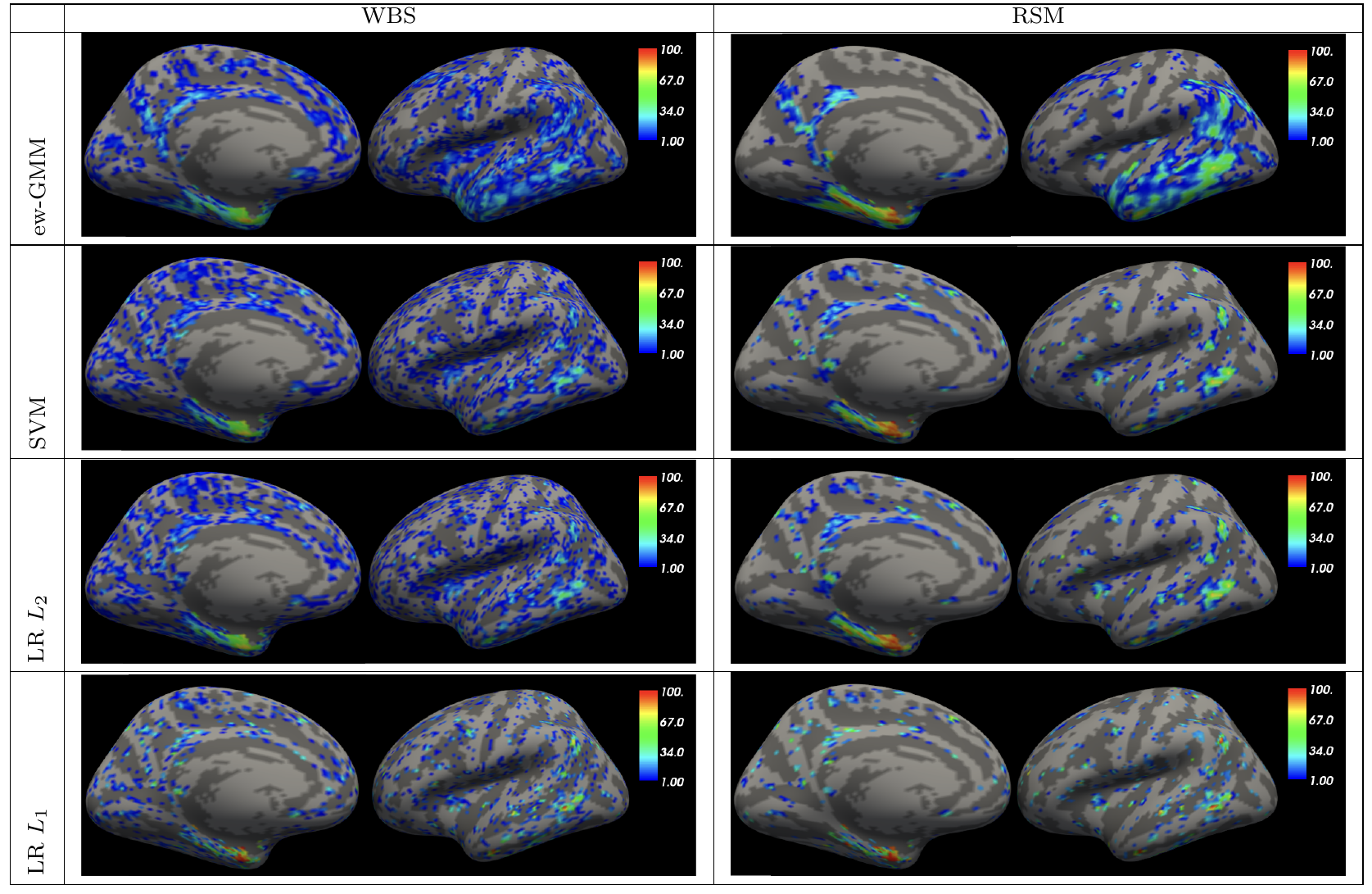}
  \end{center}
  \caption{\label{fig:frequency}{\bf Occurrence maps.} Figures show per-vertex occurrence maps for four classifiers (different rows) when bootstrap averaging (WBS) and RSM are used (different columns). At each vertex, colors indicate the number of subjects in the dataset the underlying vertex has been detected as affected by AD. Occurrence maps for bootstrap averaging are dispersed across the cortex and show high numbers around the hippocampal area. Maps for RSM are more concentrated and show higher numbers around the hippocampal area and medial temporal lobe.}
\end{figure}

Lastly, we show population-wide {\em occurrence maps} in Figure~\ref{fig:frequency}. Being able to extract subject-specific maps also allows construction of population level condition effect maps similar to conventional regression analysis. Different than the conventional maps, the values we assign to each measurement becomes more interpretable. In the maps in Figure~\ref{fig:frequency}, at each vertex we show the number of patients in the cohort that shows AD effect at the corresponding thickness measurement instead of correlation strength. That makes it possible to access  the list of subjects who showed effect at a specific vertex.  

We show occurrence maps for all the classifiers with bootstrap averaging and with using RSM. For all maps we used $l_{FPR}=0.01$ and set $\lambda=2$ for RSM. The occurrence maps from RSM were less noisy and showed high occurrence values for vertices in the entorhinal cortex and medial temporal lobe, which is consistent with the literature~\cite{dickerson2009cortical}.  The maps resulting from RSM, for ew-GMM especially, coincided with the neuropathology results of AD staging Braak and Braak presented in~\cite{braak1991neuropathological}.

Maps resulting from bootstrap averaging did not show as clearly the population-wide effects. These occurrence maps were adversely affected by the noise in individual detections. RSM's contribution to constructing cleaner subject-specific effect maps was prominent in the occurrence maps as shown in Figure~\ref{fig:frequency}.
\subsubsection{Reliability across longitudinal studies}
In this section, we discuss the reliability of subject-specific detections and advantages of RSM in this respect. Correlation scores and group differences provide one indirect evaluation of subject-specific detections. Another evaluation is the reliability of detections. If an algorithm can truly detect subject specific effects, then it should be able to repeat the detections in different images of the same subject.

To assess reliability, we used the longitudinal images in the ADNI dataset. Specifically, we used the baseline images (first images taken after screening) and images taken 6 months after. There may be differences between successive images of an individual taken 6 months apart due to various factors, e.g. normal anatomical variations, subtle changes in the acquisition devices and changes due to disease progression. Despite such factors, we assumed that differences between images that are 6 months apart should not be substantial for the ADNI dataset. Furthermore, we also assumed the same for the cortical thickness measurements extracted from these images based on reliability properties of the Freesurfer based measurements~\cite{jovicich2009mri,tustison2014large}. Consequently, we expected subject-specific detections computed from cortical thickness measurements to be repeatable between the baseline and the 6 month images in the ADNI dataset. 

Among the 290 subjects that we used previously, 268 (129 cases and 139 controls) of them had a scan acquired 6 months after the baseline, which were processed with Freesurfer without problems and passed visual quality control. We used cortical thickness measurements of these subjects. To avoid any biases, we processed the baseline and the 6 month images independently. 

We applied two independent 5-fold CV experiments on the data acquired at baseline and 6 months after. The fold definitions were kept the same in both experiments to be able to compare, however, during training and testing no information was exchanged between datasets. We quantified reliability using DSC between the detections in the baseline and the 6 months measurements of the same individual. To gather statistics, we repeated the CV experiments 10 times with random shuffling. Median values were computed over these 10 experiments for all the samples with images at both time points. 
\begin{figure}[!tb]
  \begin{center}
    \includegraphics[width=\linewidth]{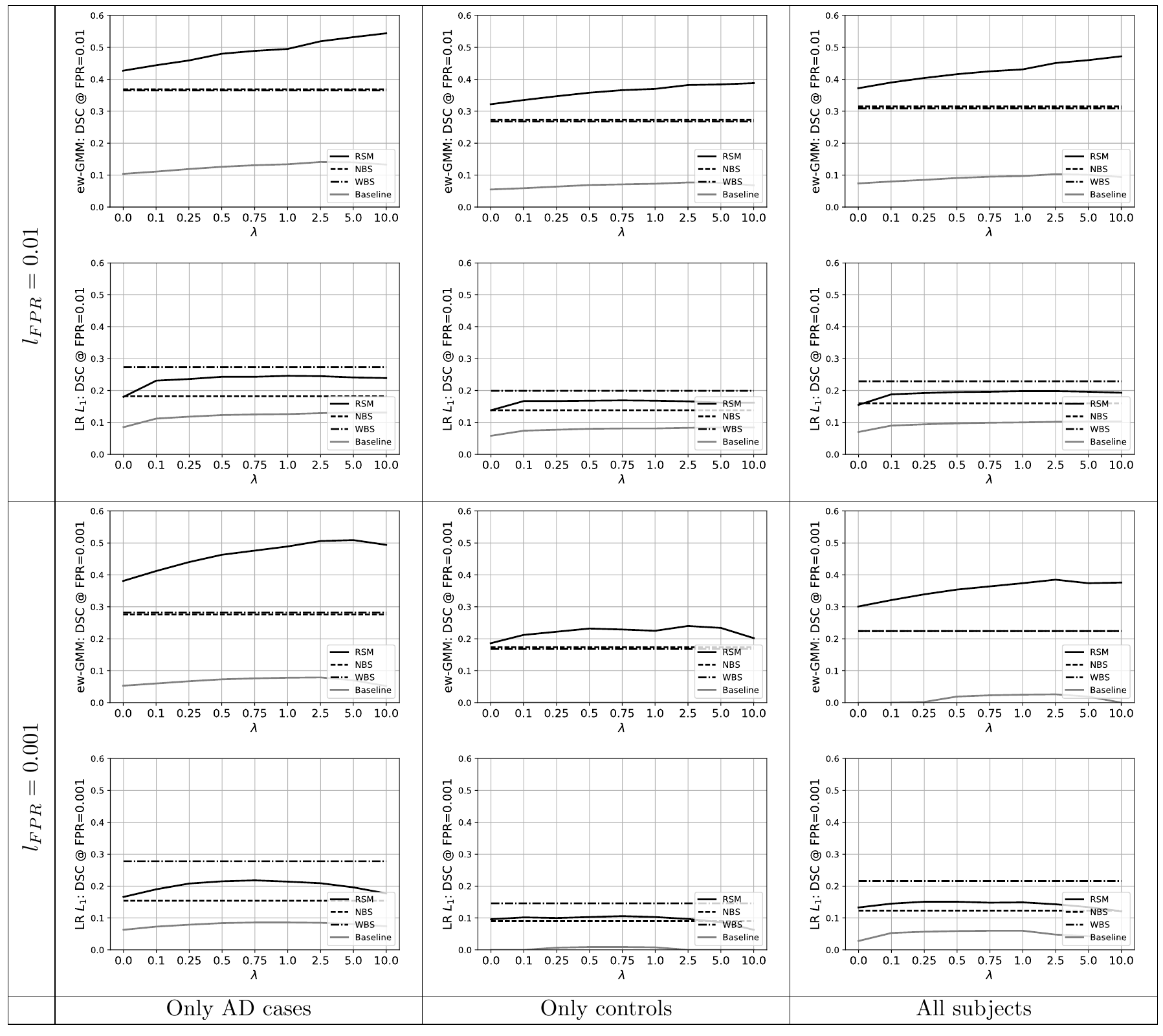}
  \end{center}
  \caption{\label{fig:reliability} \color{black}{\bf Reliability of subject-specific detections on the ADNI dataset using ew-GMM and LR $L_1$.}  Using ADNI's longitudinal dataset, we created two datasets composed of thickness maps of the same individuals extracted from images taken 6 months apart. We computed subject-specific detections in each dataset separately using ew-GMM and LR $L_1$. Graphs plot Dice's Similarity Coefficients (DSC) between detections for the same individuals in the two datasets. We show median DSC values computed over 10 randomly shuffled 5-fold CV experiments with 268 subjects. We provide three plots (different columns) for only AD cohort, only control cohort and combined. In different rows we show results for different $l_{\textrm{FPR}}$ values. In each plot we show results for using ew-GMM with no bootstrap averaging (NBS), bootstrap averaging (WBS) and using RSM with different $\lambda$ values. We also show random baseline, where DSC values were computed by comparing detections of different individuals across time. Random baselines are computed for each column using the respective cohort only.}
\end{figure}

Figure~\ref{fig:reliability} presents the quantitative reliability results for two different classifiers ew-GMM and LR $L_1$, $\lambda$ and $l_{FPR}$. The trends for SVM and LR $L_2$ were similar to that of ew-GMM so we present plots for these classifiers in the Supplementary material. The plotting are similar to the ones given in Figure~\ref{fig:aux_measures}. For each classifier, we present reliability results for detection without bootstrap averaging nor RSM (NBS), with bootstrap averaging (WBS) and using RSM with different $\lambda$ values indicated in the x-axis. We plotted results for only AD patients, only controls and all together in separate graphs. We note that although controls are supposed to be free of AD effects, this is a strong assumption and these individual might have small affected areas that can be detected repeatably with algorithms. 

Each graph shows two curves per classifier. The dark curves are median DSC and the light curves are the ``random-baseline''. We computed the random baseline as a reference by computing DSC between images of different individuals. In each plot we used the respective cohort to this end. For instance, in the AD patient case, we computed DSC scores between baseline and 6 month images of different AD patients as the reference baseline.  

Plots in Figure~\ref{fig:reliability} show several trends. First, for all plots, reliability of LR $L_1$ was substantially lower than the others. The DSC values for LR $L_1$, whether used with RSM or WBS, were lower than those of ew-GMM. This is not very surprising since the $L_1$ regularization leads to heavy dependence of the identified measurements on the training samples. Predictive set of measurements can change substantially with changing training set. Bootstrap averaging helped, however, even then the reliability was lower than the other methods in our experiments. 

RSM substantially improved the reliability of subject-specific detections for ew-GMM for both $l_{FPR}$. Improvement was substantial for detections of AD subjects.  Increasing $\lambda$ increased reliability, as can be seen in the plots. It also increased the random baseline, however, the rate of increase was much slower compared to the increase in true image pairs. For LR $L_1$, RSM had a detrimental effect compared to WBS.

For ew-GMM there was a gap between reliability of subject-specific detections and the random baseline. This gap demonstrates that the detections were indeed repeatable and hence, likely to be related to the true effects. Reliability was higher for AD patients than controls. This is not surprising since for controls we expected some level of randomness in the detections along with some true effects. The random detections were not expected to be repeatable across images. True effects on the other hand, were expected and they contributed to the reliability DSC. 
%
\subsection{Stationary vs. non-stationary pairwise term} 
\begin{figure}[!tb]
  \begin{center}
    \includegraphics[width=\linewidth]{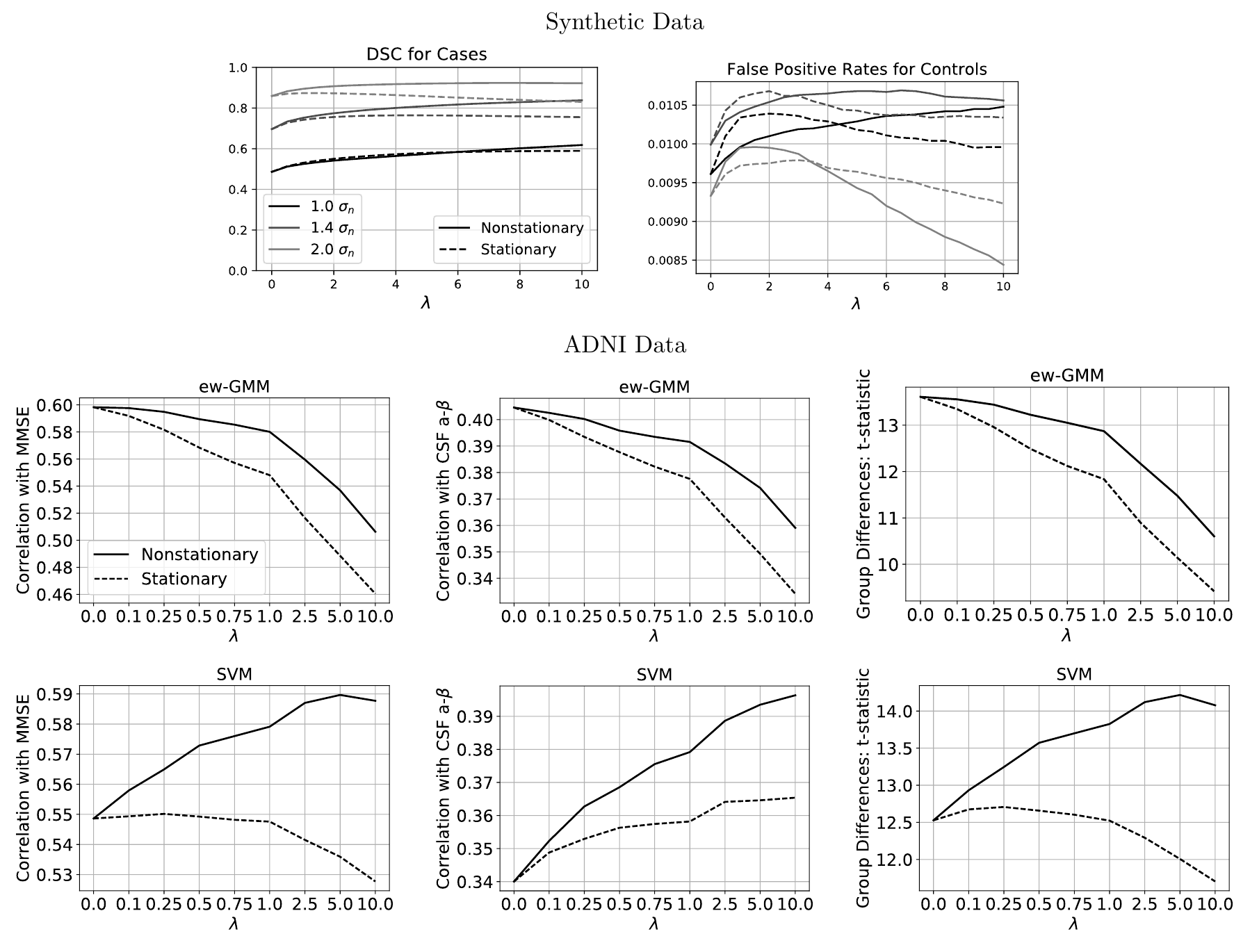}
  \end{center}
  \caption{\label{fig:stationary}{\bf Stationary vs. non-stationary pairwise term.} (Top) Graphs plot detection accuracies on the synthetic data of RSM with ew-GMM using stationary (dashed curves) and nonstationary (solid curves) pairwise terms as given in Equations~\ref{eqn:pairwise_sta} and~\ref{eqn:pairwise}, respectively. Accuracies are plotted for different $\lambda$, three effect sizes ($1\sigma_n$, $1.4\sigma_n$ and $2\sigma_n$) and $l_{\textrm{FPR}}=0.01$. We observe that the non-stationary pairwise term can achieve higher accuracies with higher $\lambda$. The trend is identical for other effect-sizes, $l_{\textrm{FPR}}=0.001$ and other classifiers, which are not shown for the sake of simplicity. \textcolor{black}{(Bottom) We show results on the ADNI dataset using different $\lambda$ values and $l_{FPR}=0.01$ for correlation scores with MMSE and CSF a-$\beta$, and group differences in detected areas. For ew-GMM, increasing $\lambda$ lead to a more rapid decrease in accuracy for stationary pairwise term. For SVM, nonstationary term lead to a substantial increase compared to the stationary term. Results for other classifiers are provided in the Supplementary materials.}}
\end{figure}

In the presented results so far, we used a nonstationary pairwise term where $\varrho_{jk}$ depended on the measurement sites. As we noted in Section~\ref{sec:method}, an alternative form could be the stationary form $\varrho$ given in Equation~\ref{eqn:pairwise_sta}. In order to better understand and empirically demonstrate the value of a nonstationary pairwise term, we compared these two alternatives on the synthetic and ADNI datasets and present results in Figure~\ref{fig:stationary}. 

For synthetic data, we present quantitative results for three effect sizes ($1\sigma_n$, $1.4\sigma_n$ and $2\sigma_n$), $l_{\textrm{FPR}}=0.01$ and the ew-GMM binary classifier. The results are presented only for these parameters and ew-GMM for the sake of simplicity. The trend was very similar for all the other cases. The nonstationary pairwise term outperformed the stationary alternative for higher $\lambda$'s. The effect was more pronounced for larger effect sizes. For smaller effect sizes the benefit of the nonstationary term became apparent at larger $\lambda$ values, while both alternatives performed very similarly for lower $\lambda$ values. This behavior supports our motivation for using the nonstationary alternative. The stationary term enforced consistency across measurements without considering any patterns in the data that suggest heterogeneities. As a result, it yielded over-smoothed and less accurate detections for higher $\lambda$ values. The nonstationary term on the other hand, captured the heterogeneities, enforced consistency only between measurements that show consistency in the training data and, as a result, achieved higher detection accuracies.

For the ADNI data, we show correlation scores and group differences for ew-GMM and SVM classifiers for $l_{FPR}=0.01$. For ew-GMM, as $\lambda$ increased the stationary term lead to a larger decrease in all the measures. This was inline with the results on the synthetic data. The nonstationary term took into account heterogeneities, did not over-smooth and lead to higher accuracy. For SVM, the positive effect of the nonstationary term was striking, it lead to substantial increase in accuracy. Results for the other classifiers and $l_{FPR}=0.001$ are presented in the Supplementary materials.
\section{Conclusions}\label{sec:conclusions}
This article proposed RSM, a reconstruction method for improving detection of subject-specific effects of a condition using binary classifiers.  Experimental results demonstrated the advantages of using RSM for different classifiers.  Detection accuracies substantially improved in the synthetic experiments.  In the analysis with ADNI dataset, RSM improved correlation between subject-specific detections and auxiliary measures for ew-GMM, SVM and LR-$L_2$.  Furthermore, reliability of detections were also higher when RSM was used.  For LR-$L_1$, RSM did not achieve improvement on detection accuracy.  However, results on synthetic data and reliability results on the ADNI dataset suggest that LR-$L_1$ might not be a good algorithm for detecting subject-specific effects.  This might be due to the $L_1$ regularization since detections are better when $L_2$ regularization is used with the same algorithm. 

In our experiments, we observed that ew-GMM with RSM yielded the best results for synthetic data and the best reliability on the ADNI dataset.  This is possibly due to the univariate nature of ew-GMM.  Multivariate methods take into account all measurements to achieve higher prediction accuracies.  In high-dimensional problems, where the number of measurements is larger than the number of samples, these methods need to use a regularization to avoid overfitting.  Regularization terms introduce additional correlation to the parameters of the model and this may be the cause of lower performance.  At this point, we should also point out that multivariate methods lack the possibility to perform localized interpretations~\cite{friston_characterizing_1995} contrary to univariate models.  In the case of subject-specific effect maps, this may also apply and might be an argument for focusing on univariate models for analysis related to subject-specific detections. 

\textcolor{black}{To the best of our knowledge, this is the first study that focuses on reconstructing subject-specific effect maps and presents a thorough analysis.  The promising results and generality of the proposed method opens up new opportunities both for applying it and improving the underlying technology.  For the applications, researchers who apply classifiers in their imaging studies can directly apply RSM and analyze resulting subject-specific effect maps.  For improving technology, we see various avenues. }

Here, we focused on linear binary classifiers due to their wide use and their interpretable nature.  Using RSM with classifiers beyond what is shown here is also possible.  However, in order to estimate parameters of the prior model and noise level, RSM uses training with bootstrap samples.  This approach might not be easy to apply on methods with computationally costly training.  Other approaches for parameter estimation may be developed for such methods.  Furthermore, here we proposed RSM for binary classifiers. Extensions to multi-label classification and regression are definitely interesting and left as future work. 

In all the presented experiments, we used one measurement per location, i.e. $f_j\in\mathbb{R}$. It is however, easy to use multiple measurements, e.g. coming from different modalities, as long as the binary classifier can utilize measurements coming from different sources. 

While determining the threshold for the continuous maps, we assumed that control subjects have no condition effects. As we pointed out earlier, this is a conservative approach since some control subjects may have condition effects. A relaxed alternative would be to assume that some percentage of the control subjects may have true effects and the threshold may be determined to limit a corresponding statistics, e.g. median of detections. If such an assumption can be made for the condition of interest then extending the proposed model is possible and would be interesting. 

Proposed model assumed that measurement error is negligible and the main source of error in predictions is variability of training sets due to sampling error.  While this assumption is valid for certain measurements, such as cortical thickness maps, for others, such as functional connectivity or quantitative MR maps, measurement errors might be large. In such cases, the proposed model can still be used but the $\sigma_j$ estimation method might underestimate this variance leading to higher false positive rates in the detections. Extension of the proposed method to handle large noise measurements is an interesting future research direction.

It is worthy to point out the low performance of the proposed method for small effect sizes. In the synthetic dataset, we show that when the effect size is low, the detection accuracies are also low. This issue is not specific to the proposed method and shared by all statistical inference methods. Nonetheless, it is still a limitation for conditions with low effect size.  Improving sensitivity of the current detections methods would be of great interest. 

RSM works with measurements that can be compared across individuals, which means it needs alignment to a common template with a registration algorithm. Registration algorithms' dependence on regularization leads to uncertainties in correspondences. Recent registration-free methods for identifying visual attributes is an interesting direction for removing the need for registration~\cite{baumgartner2017visual}. 

Lastly, we would like to point out is that subject-specific detections using machine learning tools, whether using RSM or other tools, are statistical in nature.  Detected measurements show a disease effect based on distributions observed in the training dataset.  This means detections do not necessarily indicate mechanistic alterations but only suggest them. An interesting research avenue that may improve detection of subject-specific effect maps, in this respect is to integrate such methods with biophysical mechanistic models of the disease and its progression.
\section*{Acknowledgements}\label{sec:acknowledgements}
Data collection and sharing for this project was funded by the Alzheimer's Disease Neuroimaging Initiative (ADNI) (National Institutes of Health Grant U01 AG024904) and DOD ADNI (Department of Defense award number W81XWH-12-2-0012). ADNI is funded by the National Institute on Aging, the National Institute of Biomedical Imaging and Bioengineering, and through generous contributions from the following: AbbVie, Alzheimer’s Association; Alzheimer’s Drug Discovery Foundation; Araclon Biotech; BioClinica, Inc.; Biogen; Bristol-Myers Squibb Company; CereSpir, Inc.; Cogstate; Eisai Inc.; Elan Pharmaceuticals, Inc.; Eli Lilly and Company; EuroImmun; F. Hoffmann-La Roche Ltd and its affiliated company Genentech, Inc.; Fujirebio; GE Healthcare; IXICO Ltd.; Janssen Alzheimer Immunotherapy Research \& Development, LLC.; Johnson \& Johnson Pharmaceutical Research \& Development LLC.; Lumosity; Lundbeck; Merck \& Co., Inc.; Meso Scale Diagnostics, LLC.; NeuroRx Research; Neurotrack Technologies; Novartis Pharmaceuticals Corporation; Pfizer Inc.; Piramal Imaging; Servier; Takeda Pharmaceutical Company; and Transition Therapeutics. The Canadian Institutes of Health Research is providing funds to support ADNI clinical sites in Canada. Private sector contributions are facilitated by the Foundation for the National Institutes of Health (www.fnih.org). The grantee organization is the Northern California Institute for Research and Education, and the study is coordinated by the Alzheimer’s Therapeutic Research Institute at the University of Southern California. ADNI data are disseminated by the Laboratory for Neuro Imaging at the University of Southern California.\\

\noindent We would like to thank Dr. Kilian Pohl and Dr. Isik Karahanoglu for valuable discussions. We would also like to thank the reviewers who contributed greatly to improve the quality of this work. 
\section*{References}
\bibliographystyle{abbrv}
\bibliography{cites}

\end{document}